\documentclass[a4paper, 12pt, oneside]{Thesis}  
\graphicspath{Figures/}  

\usepackage[square, numbers, comma, sort&compress]{natbib}  
\usepackage{verbatim}  
\usepackage{vector}  
\hypersetup{urlcolor=blue, colorlinks=False}  
\usepackage{setspace}
\usepackage{xcolor}
\usepackage{multirow}
\usepackage{graphicx}
\usepackage{amsfonts}
\usepackage{url}
\usepackage{hyperref}
\usepackage{float}
\usepackage{placeins}
\usepackage{amsmath}

\begin{document}
\frontmatter      

\title  {Enhancing Inference Efficiency of Large Language Models: Investigating Optimization Strategies and Architectural Innovations}
\authors  {Georgy Tyukin}
\addresses  {\groupname\\\deptname\\\univname}  
\date       {September 2023}
\subject    {}

\maketitle

\setstretch{1.3}  

\fancyhead{}  
\rhead{\thepage}  
\lhead{}  

\pagestyle{fancy}  


\setstretch{1.3}  

\acknowledgements{
\addtocontents{toc}{\vspace{1em}}  

I am grateful to my supervisor, Matt Kusner, for granting me the opportunity to work on this project and allowing me to have my own creative freedom with its direction. In addition I would like to thank  Jean Kaddour for being generous with his time and guiding my creativity to the right locations. A special thanks also goes out to Pasquale Minervini, who kindly let us use his computing resources necessary for these experiments, and without whom this project would have been left in the dust.

I would like to dedicate this work to my family and friends, for continued inspiration and support as well as for their patience with me and my sleepless nights.

}
\clearpage  

\addtotoc{Abstract}  
\abstract{
\addtocontents{toc}{\vspace{1em}}  
Large Language Models are growing in size, and we expect them to continue to do so, as larger models train quicker. However, this increase in size will severely impact inference costs. Therefore model compression is important, to retain the performance of larger models, but with a reduced cost of running them. In this thesis we explore the methods of model compression, and we empirically demonstrate that the simple method of skipping latter attention sublayers in Transformer LLMs is an effective method of model compression, as these layers prove to be redundant, whilst also being incredibly computationally expensive. We observed a 21\% speed increase in one-token generation for Llama 2 7B \cite{touvron2023llama2}, whilst surprisingly and unexpectedly improving performance over several common benchmarks.

\clearpage  


\pagestyle{fancy}  

\lhead{\emph{Contents}}  
\tableofcontents  

\lhead{\emph{List of Figures}}  
\listoffigures  

\lhead{\emph{List of Tables}}  
\listoftables  

\setstretch{1.5}  
\clearpage  
\lhead{\emph{Abbreviations}}  
\listofsymbols{ll}  
{
\textbf{LLM} & \textbf{L}arge \textbf{L}anguage \textbf{M}odels \\
\textbf{RNN} & \textbf{R}ecurrent \textbf{N}eural \textbf{N}etwork \\
\textbf{NLP} & \textbf{N}atural \textbf{L}anguage \textbf{P}rocessing \\
\textbf{MLP} & \textbf{M}ulti \textbf{L}ayer \textbf{P}erceptron \\
\textbf{ffwd} & \textbf{f}eed\textbf{f}or\textbf{w}ar\textbf{d}  \\
\textbf{ARC} & \textbf{A}bstraction and \textbf{R}easoning \textbf{C}orpus \\
\textbf{MMLU} & \textbf{M}assive \textbf{M}ultitask \textbf{L}anguage \textbf{U}nderstanding \\
}



\mainmatter	  
\pagestyle{fancy}  

\lhead{\emph{Introduction}}
\chapter{Introduction}
Large Language Models (LLMs) are becoming ever more popular in the modern world; their capabilities, whilst still not at their peak, are nevertheless impressive and will become even more so with time. With the explosive rise of ChatGPT by OpenAI earlier this year, and recent releases of Llama 2 by Meta and GPT-4 by OpenAI, there has been an increase of interest in this technology, not only in the scientific community but also by governments, corporate entities and regular members of society.

Yet LLMs are not leveraging a revolutionary technique, but rather capitalize on the wealth of curated and annotated data, coupled with the discovery of a suitable architecture, which we will cover in Chapter 2. However, the escalation in popularity has skyrocketed, due to the release of the aforementioned release of GPT-3.

In this introduction, we will expand on our motivations for the project; state the aims and objectives as well as provide the structure of the thesis ahead.

\section{Motivations and Contributions}
\subsection{Why are LLMs large, and why they will continue to grow, in parameter size}
With the public interest in large language models on the rise, there has also been an increase in creating and training better models. In particular, we can observe that the number of parameters in state-of-the-art (SOTA) LLM models is increasing exponentially each year. 

\begin{figure}[H]
    \centering
    \includegraphics[width=\textwidth]{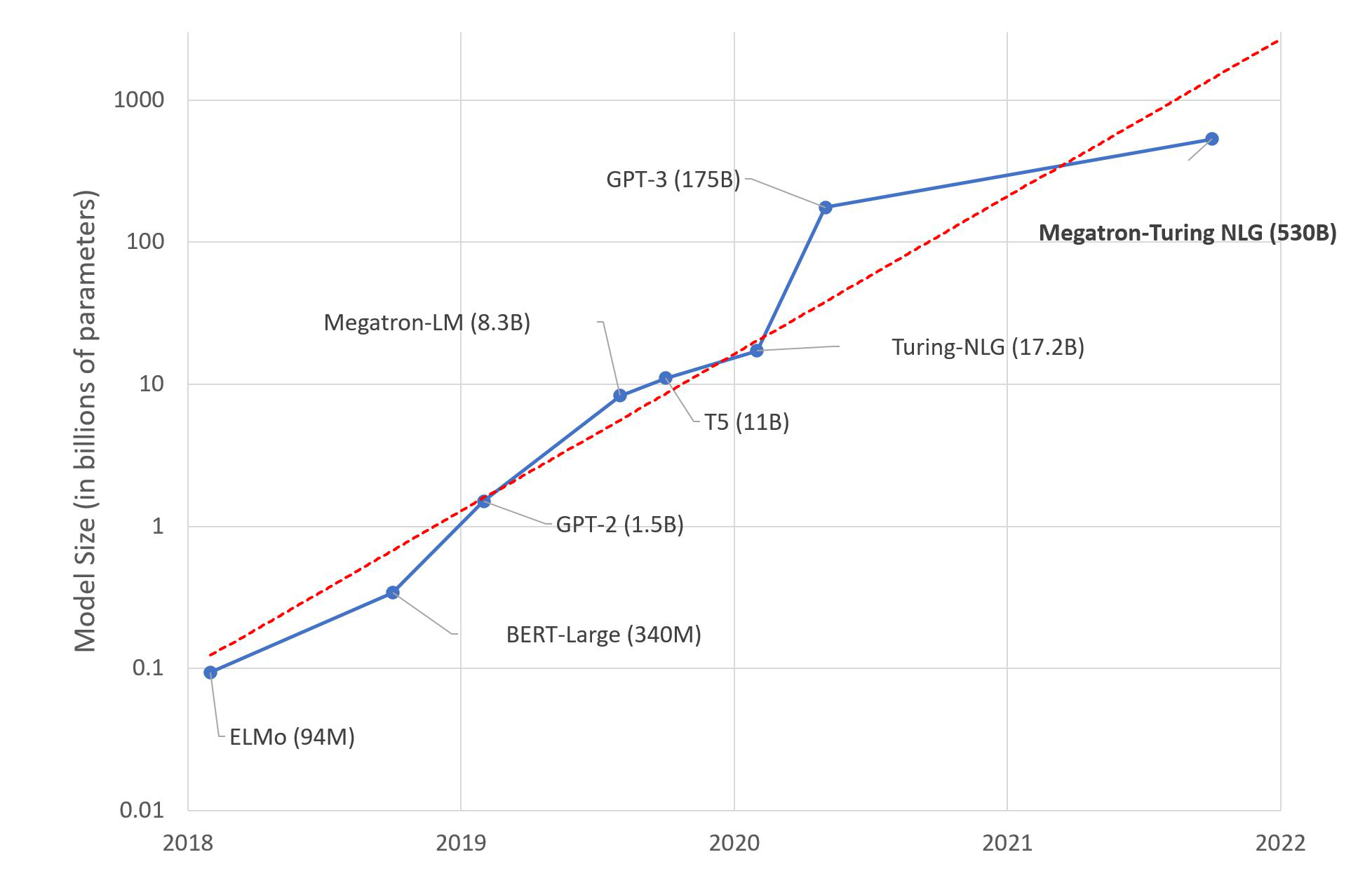}
    \caption{Logarithmic growth of model sizes over time}
    \label{fig:enter-label}
\end{figure}

This is somewhat reminiscent of Moore's Law \cite{Gustafson2011}, the observation of an exponential increase in the number of transistors in computer chips over time. Further, we notice that this increase is much faster than Moore's Law, and according to Andrew Feldman, CEO of Cerebras ``GPT-4 will have around 100 trillion parameters" \cite{WIRED}. 

A possible explanation for this increase in model size is the insatiable hunger for data. It is easy to understand that to train a better language model, we would need to feed it more data. This data is not readily available and for the most part, is also not free. In a paper by Jared Kaplan et al.  \cite{kaplan2020scaling}, they find that larger language models are significantly more sample efficient, i.e. they can be trained on significantly fewer samples than smaller models. Arora et al.  \cite{Koivisto_2018}, also empirically find that overparametrization can lead to acceleration of training of deep networks.

Does this mean that, at some point, large language models will become efficient enough to not require more data? Not necessarily. For this, let us consider a rather simple example. 
Consider a Llama 2 13B \cite{touvron2023llama2} model. We will, crudely, compare the task of predicting the next word, as a binary-classification tasks with $n$ tokens, i.e. we would like to predict to next word in a fixed sequence. Consider the simple model of predicting the correct token out of $n = 32000$ possible tokens, where Llama 2 has a vocabulary of 32000 tokens, given an input of size 4096 i.e. the size of the word embedding in Llama 2. Therefore we can model Llama 2 as a function that takes in an input of 4096 and has an output of 32000. If the final prediction unit computes a predicted token is a linear threshold unit, then by Theorem 8.9 \cite{10.5555/1795646}, we can find the Vapnik-Chervonenkis dimension, $d$, lower bound by using the number of parameters $W = 4096 \times 32000$ and the number of layers $L$, which is 2 (input layer \& output layer):

$$d \geq \lfloor \frac{W}{2}\rfloor \lfloor\frac{L}{2} \rfloor = \frac{(32000 \times 4096)}{2} \cdot 1 = 65,536,000  \times 10^8$$

Using Theorem 5.2 from \cite{10.5555/1795646}, we can bound the sample complexity necessary to achieve the desired accuracy

\begin{theorem}
    For any binary valued learning machine, with a finite VC dimension, $d$, and any distribution-agnostic learning algorithm, any $\epsilon > 0$ and $0 < \delta < \frac{1}{64}$, the size of the training set $m(\epsilon, \delta)$, with probability $1-\delta$ must satisfy 
$$m(\epsilon,\delta) \geq \frac{d}{320\epsilon^2}$$
to achieve an accuracy of $\epsilon$ away from the best possible accuracy.
\end{theorem}

If we set $\epsilon = 0.01$ (this is similar to being within 1\% of the best possible accuracy, if the risk functional is equal to the classification probability), then our required training sample size is going to be at least 2,048,000,000, and just for the last layer in the Llama 2 13B model, which is an insanely large amount of data required to ensure 
 a confident of predicting just one isolated word given a fixed sequence, therefore to be confident about all 32000 tokens we would need to scale this data by 32000. Then for another sequence we would need to do this again.
 
While there are empirical examples of larger networks being more sample-efficient, it does not prevent the increasing size of training data needed to train these networks to achieve higher accuracies. This gives birth to a vicious cycle, in which larger models are more sample efficient, but still need more data to be trained well, and we want to use that data more efficiently so even larger models get constructed.

We are already observing this in action; LLM models are becoming larger, and we expect this trend to continue. In 2018, OpenAI released its first GPT \cite{Radford2018ImprovingLU} model with 110 million parameters. Just over a year after that, NVIDIA released their MegatronLM \cite{DBLP:journals/corr/abs-1909-08053} model, with 8 billion parameters. Now it is common to see models with 40 billion parameters such as Falcon \cite{penedo2023refinedweb}; up to 65 billion in the Llama model \cite{touvron2023llama1}, (Llama 2 
 \cite{touvron2023llama2}increases this to 70 billion) and even 175 billion in GPT-3 \cite{DBLP:journals/corr/abs-2005-14165}. This increase in parameter size also increases the necessary compute for running these models, which in turn requires more GPUs and power.

The growth in required computing power and time does not just cause an inconvenience to researchers and academics as well as an increase in financial cost to users and companies. A key point is brought up by Schwartz et al. in \cite{DBLP:journals/corr/abs-1907-10597}, where they state that the expensive computations of LLMs have a surprisingly large carbon footprint. Therefore improving the efficiency of inference of LLMs is a global issue. 

\subsection{Problem we are addressing}

The above diagnostics of LLMs' size naturally lead to the question, ``Are there solutions such that we retain performance, and yet cut down on the inference costs?". This problem of reducing size whilst maintaining performance is a well-known yet open challenge. 
As previously mentioned in \cite{kaplan2020scaling}, evidence points to overparametrization having the bonus of accelerating the training, instead of expanding model capacity, however, research has not been done specifically for the case of LLMs. If the same is also true for LLMs, then it could mean that some parameters are redundant, and therefore we can remove them after training. This is further supported by Li et al. \cite{DBLP:journals/corr/abs-2002-11794}, who empirically showed that large Transformer models are more robust, compared to their smaller counterparts, when compressing after training.

Some approaches for size reduction use \emph{knowledge distillation} (use the existing model to teach a smaller model), such as in DistilBERT \cite{DBLP:journals/corr/abs-1910-01108}, whilst others suggest \emph{pruning}, or \emph{sparsification}(remove/set to 0 some of the parameters), to do this as done in SparseGPT \cite{frantar2023sparsegpt}.

Another method is, \emph{early exiting}, a technique in which layers are skipped based on whether the model is confident in its word prediction at an earlier stage. This solution has been implemented in  \cite{schuster2022confident} and  \cite{din2023jump}.

\subsection{Contributions}
The above methods all show promising results for reducing the complexity of inference, however, all these methods are complicated in nature and will take more time with larger models, due to having to efficiently pick and choose which weights are important and what you want to keep from a large model. 

In this thesis \emph{we propose a simple implementation of model compression}, which has not been explored in LLMs before. We suggest removing certain layers or sublayers without any prior analysis, in these large models could retain performance and improve time efficiency without having to perform any calculations.

We empirically show that latter attention sublayers are not critical for inference in LLMs, and even are a drawback. This research could prove useful in future inference efficiency improvements, in combination with other methods.

\section{Objectives and Hypotheses}

As previously mentioned, there are already techniques to improve the efficiency of LLM models. We will focus on `pruning': the idea of reducing the size of a neural network by selectively setting certain network components to 0 or removing them. This method has shown success in general deep networks \cite{DBLP:journals/corr/abs-2102-00554}, and in the setting where whole layers were removed \cite{backyard-dog}, as well as in LLMs \cite{frantar2023sparsegpt}.
However, \emph{sparcification} (setting certain network parameters to 0) is a computationally expensive task, and for the exponentially growing model sizes, this will become cumbersome.
Tom Brown et al. \cite{NEURIPS2020_1457c0d6} also find that LLMs are \emph{few-shot learners}, and it is not known whether reducing model size impacts this ability to learn from examples. Few-shot learning is the concept of learning a specific task, by observing some amount of examples, like humans do.
We propose a new method of pruning in the setting of Large Language Models, one that is much simpler to implement and scale up.
These lead to three hypotheses we wish to concentrate on in this thesis.

\subsection{Dropping Transformer layers}
The work by Gao Huang et al. \cite{DBLP:journals/corr/HuangSLSW16}
suggests that since earlier layers in deep learning models extract the low-level features the latter layers then rely on them, and therefore are more important. This idea is further supported by Minjia Zhang et al. \cite{DBLP:journals/corr/abs-2010-13369} in the context of accelerating the pre-training of Transformer-based models, where they implemented progressive layer dropping, to accelerate training. Their approach is stochastic: Each layer is assigned a probability of being dropped during training, which increases with time. The reasoning for this is that deeper layers don't change the overall estimation, only the quality of outputs they pass on.
\emph{We would like to investigate whether we can drop layers in already pre-trained models, and how this would affect the inference performance of LLMs.}

\subsection{Transformer's sublayers}
Each Transformer layer in a LLM can be split into two blocks, or sublayers: attention and a fully-connected multilayer perceptron (MLP) layer. Chiyuan Zhang et al. \cite{zhang2022layers} conclude that the MLP block is less robust to re-initialization and re-randomization, compared to the attention block, which points us in the direction of skipping over the MLP blocks in every layer. However, Shashanka Venkataramanan et al. \cite{venkataramanan2023skipattention} find that in Visual Transformers the attention blocks are correlated from one Transformer layer to another, resulting in unnecessary computation, and so they suggest skipping attention sublayers. \emph{We will explore how these two sublayers impact LLMs, and how skipping either one of these affects the performance and efficiency.}

\subsection{Selectively Dropping Transformer layers}
Word embedding is a technique in which separate words are then encoded as vectors, with semantically similar words being close in a vector space. Transformer blocks work by taking such a vector and outputting a vector; think of it as taking a word and outputting a word. Intuition suggests selectively skipping Transformer layers if the output vectors between consecutive layers are close. We consider this method to then decide which layers to skip, and which we cannot skip.

\section{Thesis Structure}

In the next chapter, we will first give a history of progress in NLP, then a brief overview of LLMs, the motivation for their existence, and their architectures.

In Chapter 3 we review current methods for improving LLM inference, as well as other deep learning models. We then present our methods and experiments, as well as cover the research that inspired our hypotheses

Chapter 4 then analyzes the results of these experiments, separately and then compares our pruned models against each other, in terms of performance and time.

In Chapter 5 we conclude our results and discuss the impact of our methods and future areas of research.
\lhead{\emph{Literature review}}
\chapter{Literature Review}

In this chapter, we establish how and why LLMs exist in their current form. We start off by briefly bringing up the history of Natural Language Processing, and how it has developed over the years. We then address why certain innovations were made; what issues they address, and then discuss the birth of Transformer models, the building blocks of Large Language Models. Finally, we talk about the currently dominant architecture in LLMs: Decoder-Only Transformers.

\section{Large Language Models}
Before we dive into the current LLM techniques, it is important to discuss what circumstances lead to the development of such models, what problems do Large Language models solve, and why inference is such a problem with them.
\subsection{Pre-Transformer era}
Language models are classically defined as a probability distribution over a sequence of tokens, from a finite set of possible tokens, aptly named a vocabulary, $\mathcal{V} = \{w_1,...,w_n\}$. Then each sequence of words $v_1,...,v_k \in \mathcal{V}$ can be assigned a probability of appearing, a joint distribution of words
$$p(v_1,...,v_k)$$
Then to generate a sequence, we could also sample from a joint distribution. To calculate this distribution however is its own challenge, assuming i.i.d is too naive as 
$$\text{the cat and mouse}$$
$$\text{the and mouse cat}$$
would have the same probability of being observed, even though the latter sentence does not make grammatical sense. 

It is then intuitive to have the next word in a sequence be dependent on the words before it: this is a natural decision, as this is how we understand the meaning of a sentence. As an example, consider the following sentence
$$\text{After running around he was `x' and ...}$$
The probability of the word `thirsty' compared to the word `elephant' should be different, with the former being higher.

This idea initally spawned $n$-gram language models, which captured context from the last $n$ words, followed by neural network models such as by Bengio et al. \cite{10.5555/944919.944966}, which could deal with larger context windows better.

Neural networks did manage to outperfom state-of-the-art $n$-gram models, but were much harder to train, being gatekept from advancing further by data. Thus $n$-gram language models still dominated the scene.

However, further development by Mikolov et al. \cite{RNN} \cite{5947611}, with the creation of a Recurrent Neural Network (RNN), allowed for capturing context on an infinite scale, as information was passed on from one previous nodes to the nexts, which means previous inputs were also passed on.

\begin{figure}[H]
    \centering
    \includegraphics[width=\textwidth]{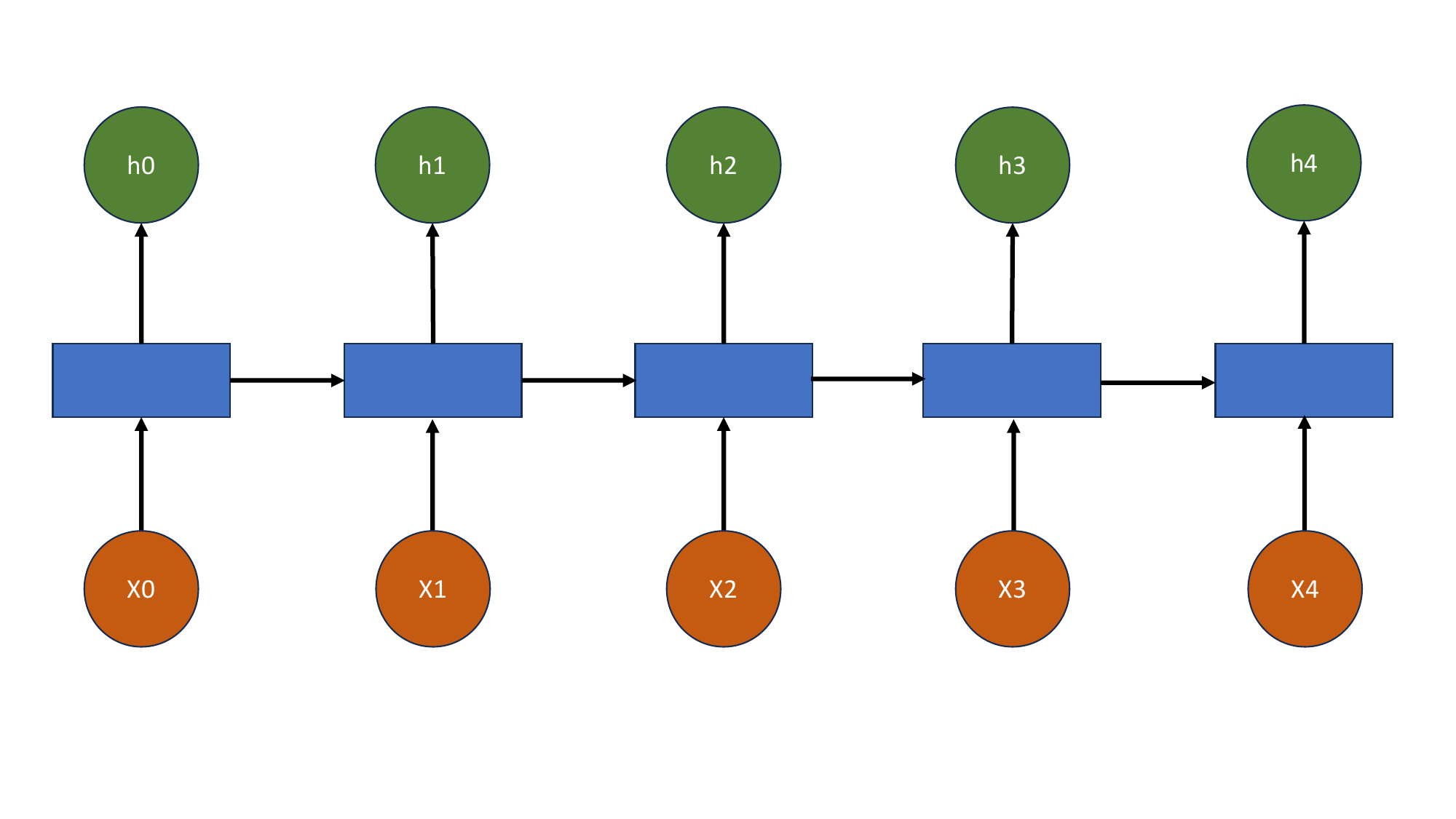}
    \caption{Unfolding of an RNN model, where the orange circles are the inputs at time $t$; the blue rectangle is a transformation applied to the current input $Xt$ and the previous output, which then outputs $ht$ and passes the same output to the next transformation in the sequence}
    \label{fig:enter-label}
\end{figure}

RNNs had a few main disadvantages which led to the innovation of Transformer-based networks.
Firstly, these were hard to train. RNN's being sequential in nature meant that standard techniques of optimization via back-propagation could not be parallelized well, and so larger models were very expensize to train. Secondly, because of this sequential nature, when back-propagation could be used it also resulted in the vanishing gradients\footnote{If the gradients are small, the multiplication of these gradients will become so small that it will be close to zero. This results in the model being unable to learn, and its behavior becomes unstable} problem, where gradients passed on to earlier weights were increasingly smaller with larger models.

\begin{figure}[H]
    \centering
    \includegraphics[width=\textwidth]{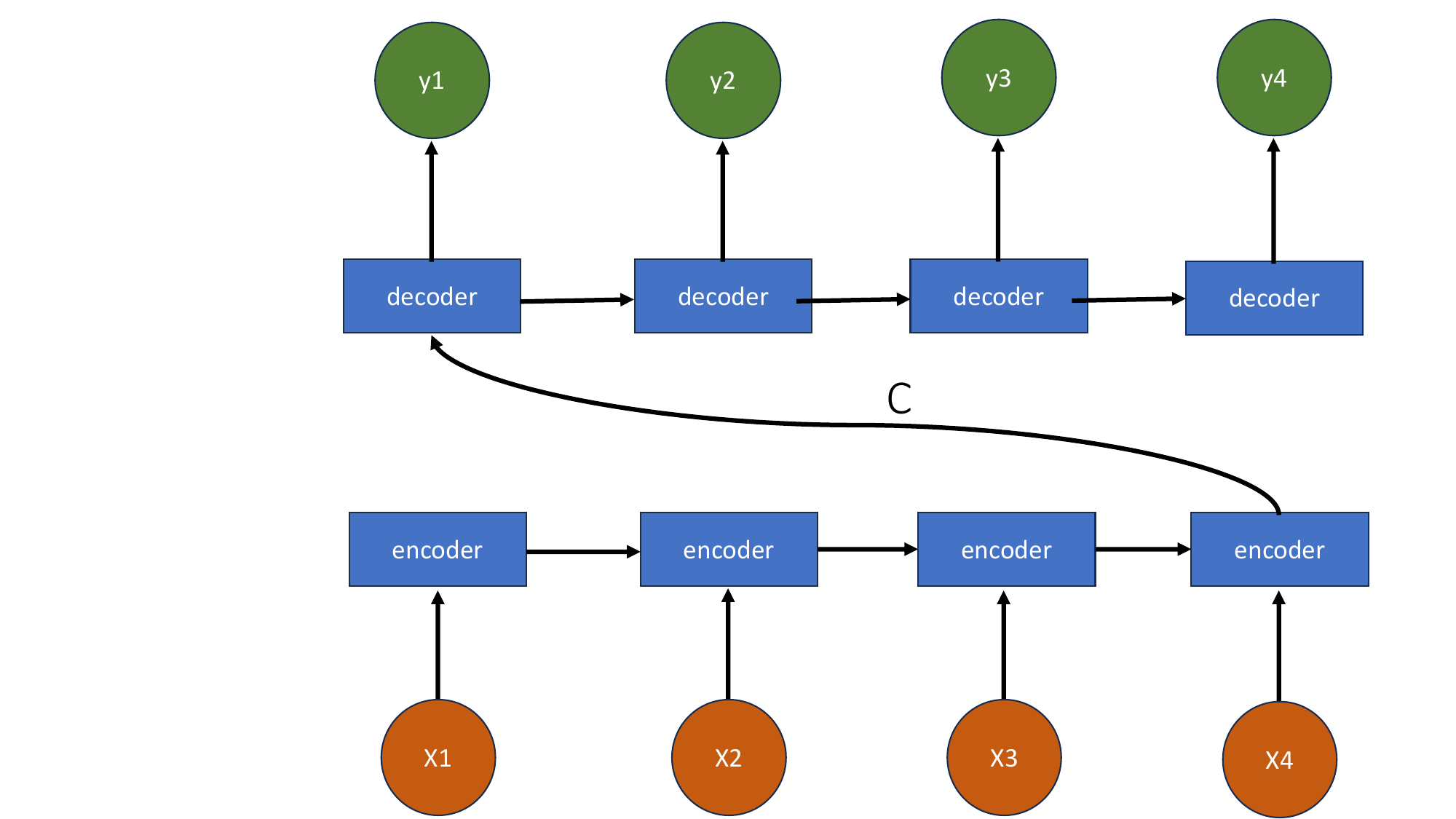}
    \caption{An example of an encoder-decoder RNN with a context vector C}
    \label{fig:enter-label}
\end{figure}

Finally, in the encoder-decoder setting, which we discuss later, RNN's communicate only the most recent output from the encoder, to the decoder. Whilst this context vector was influenced from earlier inputs, this input would be diluted, and so it limited what information the decoder could capture.

These disadvantages inspired the development of attention functions, and then inevitably Transformer models.

\subsection{Attention}
\begin{figure}[H]
    \centering
    \includegraphics[width=\textwidth]{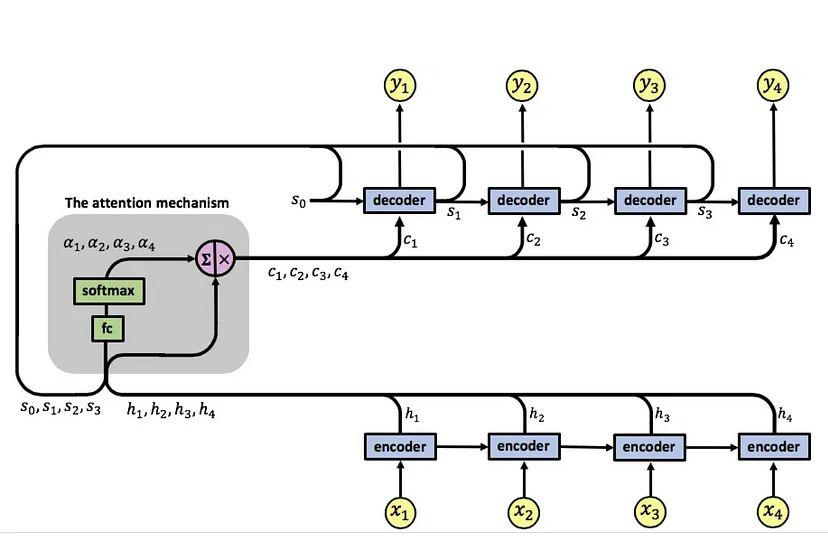}
    \caption{Standard attention mechanism for the RNN Encoder-Decoder \cite{rnn_attention}}
    \label{fig:enter-label}
\end{figure}
Bengio et al. \cite{bahdanau2016neural} first suggest using an attention layer, to capture the context of the whole sequence. Firstly, by using a bidirectional RNN (BiRNN \cite{650093}) for the encoder. It is easier to imagine this as calculating two hidden state sequences, one from a forward propagating RNN, $(\overrightarrow{h_1}, ...,\overrightarrow{h_T})$, and one for a backward propagating RNN, $(\overleftarrow{h_T},...,\overleftarrow{h_1})$.

The \emph{annotation},$h_i$, for a word $x_i$ is defined as the concatenation of the two hidden states as such $$h _i = [\overrightarrow{h_i}^T;\overleftarrow{h_i}^T]^T$$ 

These annotations are then summed to create the context vector at position $i$:
$$c_i = \sum_{j=1}^T \alpha_{ij}h_j$$

The context vector is a weighted sum of all annotations, with the weights given by a softmax transformation of the alignment model $$e_{ij} = a(s_{i-1}, h_j)$$

The alignment model scores how well the inputs (annotations) at position $j$ match with the outputs from the previous decoder position $i-1$

\begin{figure}[H]
    \centering    
    \includegraphics[width=0.3\textwidth]{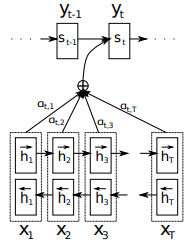}
    \caption{Attention mechanism suggested by Bengio et al. \cite{bahdanau2016neural}}
    \label{fig:enter-label}
\end{figure}

The function $a$ is a feedforward neural network, whose weights are to be also updated synchronously with the rest of the model.

Then, this context vector will be used as an input into a decoder layer $s_i$, in conjunction with the previous output from the decoder layer $s_{i-1}$.
The newly improved context layer would later be known as the attention layer, and would be used to develop the Transformer architecture.

\subsection{Encoder-Decoder}
An encoder-decoder architecture can be defined as follows:

\definition{\emph{Encoder-Decoder}}\\
Let $\mathcal{X}$ be the space of inputs, $\mathcal{Z}$ be the latent space and $\mathcal{Y}$ the space of outputs. The \emph{encoder} is defined as the function, parametrized by $\theta$ 
$$E_{\theta}: \mathcal{X} \rightarrow \mathcal{Z}$$

and the \emph{decoder}, parametrized by $\omega$
$$D_{\omega}: \mathcal{Z} \rightarrow \mathcal{Y}$$

The aim of the encoder-decoder method is to learn the latent space $\mathcal{Z}$, by updating the parameters $\theta, 
\omega$, through unsupervised learning. In an NLP setting, encoder-decoder architectures are commonly used for \emph{machine translation}, the task of translating from one language to another. The spaces $\mathcal{X}, \mathcal{Y}$ are the \emph{word-embeddings} of word vectors in language X and language Y. The latent space $\mathcal{Z}$ is the context space, which attempts to capture the context of a sequence of words, and then gets passed into the decoder, to reconstruct a similar phrase in a different language.

\subsection{Multilayer Perceptron}
A multilayer perceptron(MLP) is a feedforward neural network. Feedforward networks are neural networks in which information is only passed in one direction. 

The key distinguishing elements of an MLP is that layers of nodes are fully connected with each other, and the existence of at least 3 layers: an input layer, an output layer, and a hidden layer.

\begin{figure}[H]
    \centering
\includegraphics[width=\textwidth]{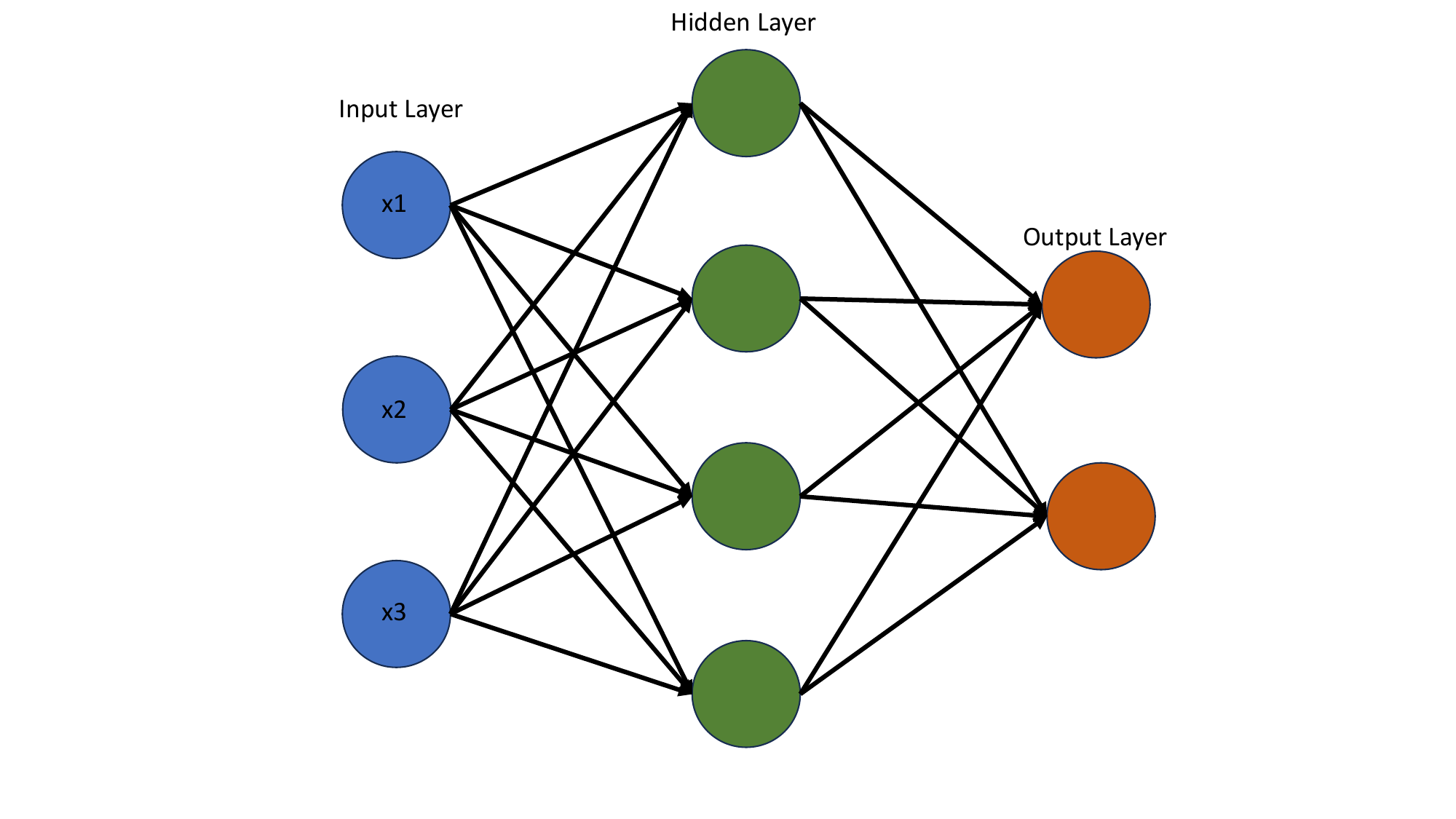}
    \caption{Example MLP structure with an input of size three, and output of size four}
    \label{fig:enter-label}
\end{figure}

Let us consider a single node in the above diagram.
The inputs is the vector $\mathbf{x} = (x_1,x_2,x_3)$.
The output of this node is then given by 
$$y = f(\mathbf{w}; \mathbf{x})$$

where $f$ is some non-linear (activation) function, and $\mathbf{w}$ is the parameter vector for the particular node. The parameter vector $\mathbf{w}$ is supposed to be learned through standard gradient backpropagation techniques.

Typically $\mathbf{x}, \mathbf{w}$ are assumed to be elements of their corresponding vector spaces (usually $\mathbb{R}^n$) and $f:\mathbb{R}^n\rightarrow\mathbb{R}$.

\section{Transformers}
The research into modern LLMs as we know them today began with the works of Ashish Vaswani et al. \cite{DBLP:journals/corr/VaswaniSPUJGKP17} when they proposed the Transformer architecture, to replace the industry standard of using Recurrent neural networks in sequence modeling.

The first Transformer architecture was built as a standard encoder-decoder structure we mentioned above, where initial embeddings are input into the encoder, which then attempts to contextualize the sequence and transfer the context to the decoder, which then outputs another sequence.

Unlike an RNN implementation of the encoder-decoder structure, which takes the sequential data one at a time, the encoder in Transformers takes the whole sequence of symbol representations $(x_1, ..., x_n)$ into a continuous representation $\mathbf{z} =(z_1, ..., z_n)$. The decoder will then take this latent representation and translate it into a sequence $(y_1, ..., y_m)$, one at a time. Because the model is \emph{auto-regressive} \cite{DBLP:journals/corr/Graves13}, the decoder takes the output at time $t-1$ as an extra input, to generate the output $y_t$. This results in sequences that can have variable length; a skill necessary for generating speech.

We will now describe the structure of this model's components: Positional Encoding, and Encoder/Decoder Structures, in which we further address the original implementation of attention in these structures.
\subsection{Positional Encoding}

First, Vaswani et al. \cite{DBLP:journals/corr/VaswaniSPUJGKP17} use positional encodings to encapsulate the importance of the position of a word $x_i$ in the sequence. The Transformer model innately lacks this ability, as it does not use any sort of recurrence or convolution on the sequence of words. 
In this particular model, positional encodings create a vector, of dimension $d_{\text{model}}$, which is the dimension of a word-embedding vector. These vectors are then added element-wise before being input into either the encoder or the decoder stack. To calculate the position encoding of a word-embedding vector in position $i$ in the sequence, $x_i = (x_i^1,...,x_i^{d_{\text{model}}})$ Vaswani et al. \cite{DBLP:journals/corr/VaswaniSPUJGKP17} suggested using 

\begin{align*}
PE_{(\text{pos},2i)} & = \text{sin}\left(\frac{\text{pos}}{10000^{2i/d_{\text{model}}}}\right) \\
PE_{(\text{pos},2i+1)} & = \text{cos}\left(\frac{\text{pos}}{10000^{2i/d_{\text{model}}}}\right) 
\end{align*}

for $0<i<\left\lfloor\frac{d_{\text{model}}}{2}\right\rfloor$. This results in a $d_{model}$-size vector with sinusoidal functions, which Vaswani et al. \cite{DBLP:journals/corr/VaswaniSPUJGKP17} hypothesize would `allow (for) the model to easily learn to attend by relative positions'.

\begin{figure}[H]
    \centering
    \includegraphics[width=0.5\textwidth]{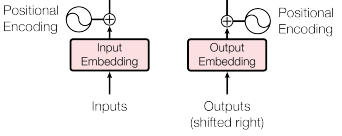}
    \caption{Application of positional encoding to embeddings \cite{DBLP:journals/corr/VaswaniSPUJGKP17}}
    \label{fig:enter-label}
\end{figure}

\subsection{Encoder-Decoder structure in Transformers}

\subsubsection{Encoder Stack}
The encoder stack consists of 6 consecutive identical encoder layers, where outputs from the layer then either go into another encoder layer or into a decoder for the last layer.

The input into the first layer is the input embeddings, where each (word-)embedded vector is added with its corresponding positional encoding vector.

\begin{figure}[H]
    \centering
    \includegraphics[width=0.3\textwidth]{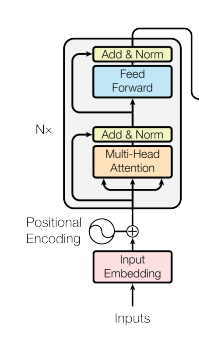}
    \caption{Encoder structure suggested by \cite{DBLP:journals/corr/VaswaniSPUJGKP17}}
    \label{fig:enter-label}
\end{figure}

Each of the layers in the encoder can be split effectively into two sublayers: the multi-head self-attention mechanism, followed by a position-wise fully-connected feed-forward network.

Around each sublayer, a residual connection is introduced. The output from a sublayer is then added with the residual, before applying layer normalization \cite{ba2016layer} i.e. the output from each sublayer will be given by $$y =\text{layernorm}(x + \text{sublayer}(x))$$

where $\text{sublayer}(x)$ is the output from the sublayer (recall, this is either the feedforward network or the multi-head attention).

What is important to note at this stage, is that to ensure compatibility with residual connections, the output of every sublayer is equal to the size of the embedding layers.

\subsubsection{Layer Normalization}
Layer normalization \cite{ba2016layer} is a technique, which was developed to reduce training times, by keeping outputs normalized. Normalization is not a novel concept, with batch-normalization existing \cite{DBLP:journals/corr/IoffeS15}
since 2015, and giving good results for accelerating the training of standard feedforward networks.

In deep learning, gradients of weights in one layer are reliant on outputs from a previous layer. Batch-normalization was introduced to combat this overall dependence, by rescaling the outputs from the previous layer, using the empirical mean and standard deviation of the mini-batch. Thus, to normalize some outputs $x_1, ..., x_m$ from a linear vector space, in a mini-batch, we calculate the mini-batch mean $$\mu_B = \frac{1}{m}\sum_{i=1}^mx_i$$ and mini-batch standard deviation $$\sigma_B = \sqrt{\frac{1}{m}\sum_{i=1}^m(x_i - \mu_B)^2}$$

These values are then used to normalize the outputs $$\hat{x}_i = \frac{x_i - \mu_B}{\sigma_B}$$.

The flaw with this method is that a batch-normalization layer then is dependent on the mini-batch size, which proves to be an issue when we want a varied input size, such as in recurrent networks and Transformers.

Layer-normalization avoids this by not considering the mini-batch size, but rather the number of output nodes from the previous layer, so if some layer $L$ has 5 output nodes, which produce some outputs $\{x_1,...,x_5\}$, then $m$ gets replaced by 5 in the above equations.
The advantage of this approach is most evident in recurrent neural networks, and after, obviously in the Transformer models, since mini-batch sizes might vary per sequence, both in training and testing and so having a normalization layer dependent on the model rather than the mini-batch size seems more appropriate, and was empirically shown to work better.

\subsubsection{Decoder Stack}

\begin{figure}
    \centering
    \includegraphics[width=0.25
    \textwidth]{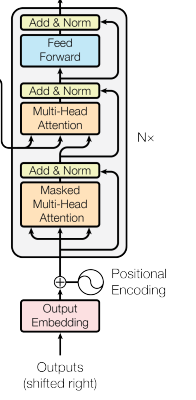}
    \caption{Decoder structure suggested by \cite{DBLP:journals/corr/VaswaniSPUJGKP17}}
    \label{fig:enter-label}
\end{figure}

The Decoder stack consists also of 6 consecutive identical decoder layers. The decoder layer takes the output embedding of the previous decoder layer, with positional encoding added, as well as the output from the encoder.

The main difference of this decoder layer is the addition of another masked attention multi-head attention sublayer (with a residual connection and layer normalization), which intakes the previous decoder outputs and applies a `masking', to make sure that predictions for a word at position $i$, can only be made from words in earlier positions smaller than $i$.

\subsubsection{Multi-head attention}
The most significant part of this Transformers model is the multi-headed attention. Vaswani et al. \cite{DBLP:journals/corr/VaswaniSPUJGKP17} introduce a \emph{Scaled Dot-Product Attention}, which takes three vectors $q,k$, of dimension $d_k$, and $v$, of dimension $d_v$.

\begin{figure}[H]
    \centering
    \includegraphics[width=0.4\textwidth]{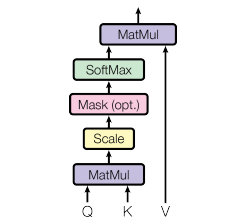}
    \caption{Scaled dot-product attention structure \cite{DBLP:journals/corr/VaswaniSPUJGKP17}}
    \label{fig:enter-label}
\end{figure}

Since we are trying to capture the attention of word-embedding vectors, it makes sense to stack these embeddings into matrices i.e. $Q \in \mathbb{R}^{d_k \times d_{\text{model}}}, K \in \mathbb{R}^{d_k \times d_{\text{model}}}, V \in \mathbb{R}^{d_v \times d_{\text{model}}}$ where $d_\text{model}$ is the size of a single word-embedding, where $d_v, d_k$ can be set by the user.

\begin{figure}[H]
    \centering
    \includegraphics[width=0.4\textwidth]{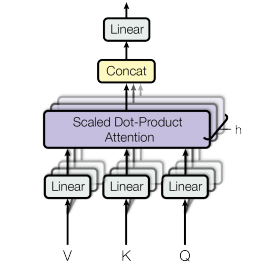}
    \caption{Multi-head attention structure}
    \label{fig:enter-label}
\end{figure}

Then the attention function, for given matrices $Q, K, V$ is given as:

$$\text{Attention}(Q,K,V) = \text{softmax}\left(\frac{QK^T}{\sqrt{d_k}}\right)V$$

The factor of $\frac{1}{\sqrt{d_k}}$ prevents softmax gradients from becoming extremely small. To show that this can be an issue, imagine two independent random variables $a,b \in \mathbb{R}^n$, with mean 0 and variance 1. Their dot product is then 
$$a \cdot b = \sum_{i=1}^n a_i b_i$$ which has a mean of still 0, but the variance is now $n$. This factor $\frac{1}{\sqrt{d_k}}$ normalizes the matrix multiplication, to prevent it from growing, which would push smaller gradients even smaller. 

Vaswani et al. \cite{DBLP:journals/corr/VaswaniSPUJGKP17} found that projecting the vectors $q, k, v$ to $h$ different learned linear projections, and then performing the Scaled Dot-product attention on the projected matrices improves training, as it allows the process of calculating the Multi-head attention output via parallelization of the $h$ stacks.

For the self-attention (attention function which takes inputs only from the layer it is in) both the encoder and decoder take outputs from the previous layers for all three matrices $Q,K,V$. The decoder also makes use of the masking layer which sets some inputs to $-\infty$. This is done to allow the model to be auto-regressive, and continue to make predictions for the next word, only using the ones it previously generated.

We also notice the secondary attention function in the decoder layer, which takes the input from the encoder as matrices $K,V$. This allows the decoder to utilize the captured attention from the whole sequence. The $Q$ is then passed in from the previous (masked) attention function in the same layer.

\subsubsection{Feedforward Network}
The feedforward network in the Transformer is rather simple. It consists of an input layer, of size $d_{model}$, a hidden layer $h$ of any size $d$, and an output layer of size $d_{model}$ again.

The hidden layer performs an affine transformation on the input $x$, followed by a ReLU (Rectified Linear Unit)\cite{4082265}, to add non-linearity to the data.
$$f(x) = \text{max}(0,xW_1 + b_1)$$

Then another affine transformation is used to create the outputs

$$y = f(x)W_2 + b_2$$.

$W_1, W_2, b_1, b_2$ are weights to be learned.

\subsubsection{Summary}
In the above section, we summarized the motives for creating the Transformer architecture, as well as describing the reasons as to why certain decisions were made when creating this kind of model.

We also briefly touched on what makes Transformer architecture, and the key point of not having to use recurrence, within the model, allowing for parallelization of training, in other words, training such models becomes more time-efficient at the cost of computational complexity (since the context is being captured from the whole sequence). 

This means that longer sequences are more costly to interpret using this Transformer setting, compared to Recurrent Neural Networks. A solution is presented in the paper, to limit the size of the context window, but it is exactly this issue that sparks the next, and the most current, innovation into Transformer architecture.

\section{Decoder-only Transformers}
The first mention of a decoder-only architecture was mentioned by Liu et al. \cite{DBLP:journals/corr/abs-1801-10198}, for the task of summarizing long sequences of texts. This is exactly the setting in which the encoder-decoder architecture would struggle in terms of complexity, due to having to capture the context for the entire sequence with the encoder.

Liu et al. \cite{DBLP:journals/corr/abs-1801-10198} propose a radical, yet simple solution: completely omitting the encoder from the Transformer model proposed earlier. As a bonus, this setup allows for almost 50\% of parameters to be dropped. Liu et al. \cite{DBLP:journals/corr/abs-1801-10198} also suggest that in monolingual text-to-text tasks, redundant information is re-learned in both the encoder and decoder, and so removing the encoder, allows for easier optimization, which they then proved empirically.

\subsection{Main Changes}
After removing the encoder module, the decoder-only Transformer is then trained as a standard language model, as the main objective is to predict the next word, given the previous words. For this, a simple transformation needs to be done to the inputs and outputs. Rather than create a function that reads inputs $(x_1,...,x_n)$ and generates outputs $(y_1, ..., y_m)$, Liu et al. \cite{DBLP:journals/corr/abs-1801-10198} convert \emph{both} inputs and outputs into a single sentence 
$$(w_1, ..., w_{n+m+1}) = (x_1,...,x_n,\delta,y_1,...,y_m) $$
where $\delta$ is a special character that acts as a separator.

The goal of this Transformer is to then predict the next word, $y_{m+1}$

$$p(w_1,...,w_{n+m+2}) = \prod_{i=1}^{n+m+1} p(w_i|w_1,...,w_{i-1})$$

This Transformer model can be further improved by using updated attention functions, to address the complexity of the attention functions introduced by Vaswani et al. \cite{DBLP:journals/corr/VaswaniSPUJGKP17} 
\emph{Local attention} functions divide the sequence into regular blocks of words and multi-headed attention is performed on all those blocks, in parallel. This implies that by fixing the size of a block, the computational complexity of an attention function will be constant, and therefore for longer sequences the complexity is only $\mathcal{O}(n)$ \footnote{Big O notation refers to the highest order of complexity of an algorithm i.e. how it scales with some parameter $n$. In this case we mean to say that our complexity is linear, which means that for large enough $n$ we expect doubling the $n$ as increasing complexity by a factor of 2}, compared to the quadratic growth initially described by Vaswani et al. \cite{DBLP:journals/corr/VaswaniSPUJGKP17}

\begin{figure}[H]
    \centering
    \includegraphics[width=0.25\textwidth]{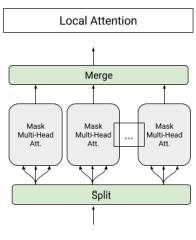}
    \caption{Local attention \cite{DBLP:journals/corr/abs-1801-10198}}
    \label{fig:enter-label}
\end{figure}

The next change was the \emph{Memory-compressed attention}. By adding a strided-convolution for the matrices $K,V$, it reduces the number of operations performed in the dot-product, in the Scaled Dot-Product attention function, whilst also obtaining global context from the sequence, which the Local Attention is not capable of doing. 

\begin{figure}[H]
    \centering
    \includegraphics[width=0.28\textwidth]{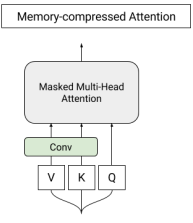}
    \caption{Memory Compressed attention \cite{DBLP:journals/corr/abs-1801-10198}}
    \label{fig:enter-label}
\end{figure}

\section{Current Research}

The topic of improving efficiency of large language models, as well as other deep neural networks, is an increasingly important field. Below we will cover some current research into existing approaches as well as relevant research into our field of interest.

\subsection{Knowledge distillation}

Knowledge distillation was first introduced as a method of model size reduction in DistilBERT. \cite{DBLP:journals/corr/abs-2002-11794}

We first specify a reduced model size, that we would like to compress our original model into. Our aim with knowledge distillation is to train a smaller (student) model, using the knowledge acquired from the larger original (teacher) model. This then transforms the task from the usual unsupervised learning, of learning the word patterns, into a supervised learning task, with the objective usually being a combination of learning the target distribution for the next word, given by the original model; trying to match the hidden outputs between the student and teacher using \emph{cosine similarity}, as well as any other loss that might seem acceptable in the situation.

For example, since BERT \cite{DBLP:journals/corr/abs-1810-04805} used a Masked Modeling Loss\footnote{This is the loss derived from the probability distribution of the predicted words compared to the correct word.} during training, it made sense that DistilBERT should also try to minimise this loss.

\subsection{Pruning/Sparcification}

Model reduction can also be achieved by removing weights that are redundant, or by setting these to 0. The most well known algorithm for sparcification in machine learning is the L1 LASSO regularisation \cite{rs15061670}, which aims to set extremely small weights to, thus reducing computation complexity, as operations involving 0 are efficient.
This method was identified as an effective technique \cite{NIPS1989_6c9882bb} \cite{NIPS1992_303ed4c6} \cite{han2015learning} \cite{molchanov2017pruning} \cite{li2017pruning} to improve the efficiency of deep networks for applications with limited computational budget. The typical procedure of implementing pruning is to first train an overparametrized model, for reasons mentioned before [citation needed here], prune the weights of the model before finally fine-tuning the newer smaller model, to compensate the reduced size.
The reason for training large and then decreasing size is that these types of models outperform smaller models \cite{li2017pruning} \cite{luo2017thinet}. The models are then fine-tuned \cite{han2015learning}, as the remaining architecture and weights are thought to be essential to achieve the optimal model. Network pruning can be split into two main categories: individual weight and structured pruning.
As the names suggest, individual weight pruning focuses on optimizing individual parameters, whereas structured pruning attacks whole layers or input channels, i.e. the architecture of the network.

Liu et al. 2019 \cite{DBLP:journals/corr/abs-1810-05270} investigate the differences of individual weight pruning and structured pruning, and conclude that individual weight pruning could potentially trap the model into a local minimum, due to being forced to stick with the same architecture.
Structured pruning avoids this issue, as the goal is to find the most optimal reduced architecture, and then the remaining parameters are updated accordingly.
Another advantage of using structured pruning, is that while individual pruning can reduce computational complexity, the sparse weight matrices do not instantly yield results in efficiency without the dedicated hardware/libraries \cite{han2016eie}
However Zhu \& Gupta \cite{zhu2017prune} suggest that large-sparse models outperform small dense models, and so there is no rule, as to what pruning is the best.

Skipping full layers has also existed as a concept, but in the case of general deep neural networks. It was presented in the works of Gorban et al. \cite{backyard-dog}, where they adapted deep neural networks such as the VGG-16 \cite{simonyan2015deep} model for face recognition into a shallower model. This approach has not been used for Large Language Models yet.

An interesting idea is also presented in the works of \cite{NIPS2017_a51fb975} in regards of pruning the structure of the network in accordance to the current inputs into the network, which is similar in fashion to comparing the student and teacher hidden states in knowledge distillation.

\subsection{Quantization}
Quantization, in the context of model compression, refers to reducing the number of bits used to store weights and parameters in models. For example, reducing the weights from a 16-bit floating point number, to an 8-bit floating point.
Quantization is therefore an important technique, that is needed to be able to load large models onto computers with low capability. It has been shown that moving from a floating point representation to low-precision, fixed integer values, such as 4 bit representations, can successfully reduce the required memory costs as well as the inference time \cite{gholami2021survey}. Gholami et al. \cite{gholami2021survey} show that this speed up can be upto a factor of 16.

The obvious downside, is the fact that accuracy will get impacted however, since we are dropping in precision. Most common methods of performing quantization is done through rounding or truncating high precision values to low precision, however there are also other methods of mapping from the high precision values to lower ones.

\subsection{Early Exiting}
Early exiting can be thought of as an adaptive variation of structured pruning. If the network already knows what the output should be at at earlier layer, then the following layers are redundant in producing the outputs. This follows from the works of Zhang et al. \cite{zhang2022layers}, who observed that in layer-wise, overparametrized models (such as LLMs), earlier layers are more sensitive to reinitizialization, which implies that these layers hold most of the information in deciding on a specific output.

\begin{figure}[H]
    \centering
    \includegraphics[width=0.5\textwidth]{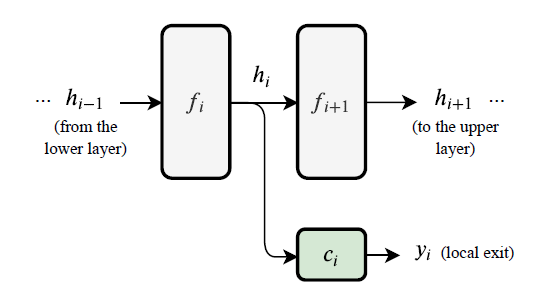}
    \caption{Early Exit example \cite{early_exit}}
    \label{fig:enter-label}
\end{figure}

Early exiting was first introduced by Teerapittayanon et al. \cite{teerapittayanon2017branchynet} as a solution to the growing parameter size of deep neural networks, in which multiple exit branches were introduced into the network architecture. These would assess the current hidden states, and given a threshold value, would exit the inference stage early, if the entropy of the output was high enough, i.e. the model was confident enough in its decision.

There are few issues with this approach, however. If we were to exit at the optimal point for each input, this would require the model to perform this check after every layer, considerably increasing the computational cost of inputs, where the model is not confident enough to exit early. This issue is further amplified if the intermediate outputs need to be preprocessed into the probability distribution over all labels.

\section{Summary}

To summarize this chapter, we covered the history of Natural Language Processing, how this field developed, and how Large Language Models fit into this area of machine learning. Afterward, we discussed what problems Large Language Models solve, and why it is necessary to have a specific Transformer structure.
We covered key building blocks in LLM architecture, which we need to understand before attempting to reduce model size.

In the next chapter, we show current research in the problem of optimizing the \emph{inference} of LLMs, and the intuition behind our methods and experiments. 
\lhead{\emph{Methods}}
\chapter{Methods}

In this chapter, we first restate the problem we are trying to solve. We then dive into previous research conducted into the optimization of deep learning models and Large Language models; and discuss the drawbacks and advantages each one provides.
Next, we cover our proposed method, its novelty, as well as the hypotheses we aim to test. Finally, we consider the setup of the experiments.

\section{Inspiration}

As we previously stated, the problem of reducing model size is extremely important for the sake of utilizing all the bonuses of large language models, and we predict that this will be a gating factor of larger models. The solutions we discussed above have their merits and have been shown to provide impressive reductions in parameter count, with little hindrance to performance, when compared to their full-size counterparts.

However, all the suggestions above require complicated processes: whether this be the retraining of the student model, as in knowledge distillation, or meticulously calculating which weights to exclude in a given layer, such as in SparseGPT \cite{frantar2023sparsegpt}.

We want to answer the question “Is there a simpler method, that can still provide good results in performance?”

Inspired by the conclusions drawn by Gao et al. \cite{DBLP:journals/corr/HuangSLSW16} and Gorban et al. \cite{backyard-dog} we suggest it could be sufficient to drop \emph{full} Transformer layers in a Large Language Model. Dropping layers has been used by \cite{DBLP:journals/corr/abs-2010-13369}, but this was in the context of pretraining a model. Our goal is slightly different, we would like to apply this method on a pre-trained LLM, without having to retrain the weights fully, limiting ourselves to only fine-tuning and few-shot learning on specific tasks.

An alternative to the rather radical method of dropping full layers is to only look at sublayers in a Transformer layer, namely the attention sublayer and the feedforward network layer. \cite{zhang2022layers} discusses that the sublayers in Transformer layers are sensitive to changes differently. For example, attention layers are more robust (less sensitive to a perturbation to their weights) than MLP. 

The works of Venkkataramanan \cite{venkataramanan2023skipattention} further inspires exploring this method for LLMs. In \cite{venkataramanan2023skipattention}, latter attention sublayers were skipped in Visual Transformer models, and it was shown that skipping these computationally expensive layers \cite{DBLP:journals/corr/VaswaniSPUJGKP17} sped up inference. Recall that attention sublayers attempt to capture context from the whole sequence in Transformer layers \cite{DBLP:journals/corr/VaswaniSPUJGKP17}. This means that they have a computational complexity of $\mathcal{O}(n^2)$ for sequences of length $n$. We predict that skipping attention sublayers will result in better performance than skipping the feedforward sublayer when comparing the two as methods for reducing model size.

Lastly, we would like to explore whether dropping layers selectively rather than just the last ones, will improve performance. Whilst one can try all combinations of skipping layers, we suggest that using the cosine similarity between the layers is sufficient to distinguish important layers from redundant ones. Recall that in a Transformer layer \cite{DBLP:journals/corr/VaswaniSPUJGKP17} the output and input of the layer are designed to be of the same dimension. Then we suggest that layers that have a high cosine similarity between the input and output are redundant, as they neither capture nor instill a significant amount of information for the next layer. Cosine similarity, as previously discussed, formed part of the training objective in DistilBERT \cite{DBLP:journals/corr/abs-1910-01108} knowledge distillation, where the aim was to have the hidden states of the teacher model and the student model correlated.

We propose that cosine similarity alone can be enough to judge whether a layer is significant or not.

\section{Cosine similarities}
Recall that word embeddings are vector representations of words, or `tokens', from a vocabulary, with the idea that words that semantically close together, will be close together in the embedding vector space.

Cosine similarity is a measure of how close two vectors are, based on the angle between them. Vectors that are parallel are assigned a score of 1, (angle between them is 0, and $\cos({0}) = 1$), whilst perpendicular vectors are given a score of 0 (angle between them is $\frac{\pi}{2}$).

Cosine similarity for two vectors, $v,w \in \mathbb{R}^n$ is given by
$$\text{score} = \frac{v \cdot w}{\|v\|\|w\|}$$

In the context of LLMs, we can apply cosine similarities on the outputs between layers, and compare how much a layer changes the word embedding. If a cosine similarity is high, then that layer does not instill much information, and vice versa.

\section{Experimental setup}

To conduct our experiments, we will use the lm-evaluation-harness developed by EleutherAI\cite{eval-harness}, and evaluate our compressed models against popular LLM benchmarks.

The harness developed by EleutherAI is a common way of analyzing various metrics for given pre-trained models. It is the go-to method of evaluating models for the OpenLLM leaderboard \cite{open-llm-leaderboard}, a leaderboard for various LLM models, all evaluated on the same benchmarks. We will be using these benchmarks as well to evaluate how much of a performance decrease we observe when we start skipping layers, bar one benchmark which we discuss the reasons for later.

We decided to use Llama-2 \cite{touvron2023llama2} for our experiments on skipping layers, and we chose to work on Llama-2 7B and 13B, to evaluate how different sizes are affected by our changes. We stuck with one model, for the reason of consistency, however this obviously also has its drawbacks.

To accommodate for skipping layers, we had to amend the EleutherAI \cite{eval-harness} harness, as well as make changes to the Llama-2 model in the transformers package in Python, which we provide here : 

\url{https://github.com/Geeeorgy/lm-evaluation-harness}.

If you would like to reproduce our results, then you will have to transfer the 
\texttt{modeling\_llama.py} file into the transformers package folder.

\subsection{Benchmarks}

OpenLLM Leaderboard uses four benchmarks, which was defined by the community, as the most optimal way of deciding the performance of a large language model.

These are \emph{ARC challenge, Hellaswag, TruthfulQA, MMLU}, each one focusing on the model’s ability to tackle various language tasks it could be expected to perform.

Each of the above experiments has an optimal fewshot number, which we also apply to each of your experiment runs, thus also investigating how well these compressed models are able to learn of examples

Below, we give a description of our experiments on the Llama models. It is worth noting that Llama 2 7B has a total of 32 layers, and Llama 2 13B has a total of 40 layers. Then skipping 10\%, 25\%, and 33\% is the equivalent of skipping 3, 8, and 11 layers for Llama 7B (4, 10, 14 for Llama 2 13B) respectively.

\subsubsection{ARC challenge}
Abstraction and Reasoning Corpus(ARC) \cite{clark2018think} serves as a testing ground for evaluating a neural networks skill acquisition in tackling unfamiliar tasks. It challenges neural networks to learn complex tasks, having only seen a few examples (few-shot learning), offering a glimpse into a future where neural networks can rapidly adapt to solve new problems independently.

For Large Language Models, the ARC dataset is a collection of questions, aiming to test the ``knowledge" that an LLM possesses. For example,
\begin{verbatim}
    "A fold observed in layers of sedimentary
    rock most likely resulted from the"
\end{verbatim}
It is then the job of the LLM to correctly predict the next sequence, given multiple choices, such as:\
\begin{verbatim}
    A. "cooling of flowing magma."
    B. "converging of crustal plates."
    C. "deposition of river sediments."
    D. "solution of carbonate minerals."
\end{verbatim}
There is only one correct answer for every question, in this example this is B. The standard metric for measuring how well an LLM network performs is normalized accuracy. This is better than regular accuracy, in terms of understanding how well an algorithm is tackling different types of questions. Consider the situation where most questions have the answer `B' as the desired answer. Then an algorithm that predicts `B' for every question would score a high accuracy on the example set, but will perform worse on sets that do not have `B' as the correct answer. Therefore normalized accuracy is used, to avoid having such issues.

\subsubsection{Hellaswag}
HellaSwag \cite{zellers2019hellaswag} is another dataset that attempts to measure the knowledge of a Large Language model. HellaSwag introduces works in a similar fashion to ARC, where given a sequence, usually a full sentence, the LLM is tasked with predicting the next sentence, given a few options.
The main selling point of this dataset is these example sequences, where incorrect answers do not make sense to humans, however are misclassified by state-of-the-art models.
Consider the following prompt:
\begin{verbatim}
    "A man is being pulled on a water ski as he
    floats in the water casually. he
\end{verbatim}
The large large language model is then given the answers:
\begin{verbatim}
     1. "mounts the water ski and tears through the water at 
                fast speeds."
     2. "goes over several speeds, trying to stay upright."
     3. "struggles a little bit as he talks about it."
     4. "is seated in a boat with three other people."
\end{verbatim}
where 3 is the correct answer. The other answers, do not make much sense for us, but for language models these answers do look similar.
Again, the metric used to test LLMs is normalized accuracy, for reasons stated above.
\subsubsection{TruthfulQA}
Large language models are known to be capable of generating false statements \cite{shuster2021retrieval} \cite{krishna2021hurdles}, and so it is important that we measure how truthful a model is. 
TruthfulQA \cite{lin2022truthfulqa} does this by measuring how well LLMs can identify truthful answers to questions. We use Multiple-Choice 2 (mc2) as the metric for measuring this benchmark. The LLM is provided a collection of both truthful and false answers. The score is a normalized probability distribution assigned to truthful answers.

However, TruthfulQA shows that larger models are indeed less truthful, and this is an expected result if the model is trained on false data. Consider a model that is trained on data from online forums. Whilst forums might be able to convey the structure of sentences, they can also express untruthful answers, especially when people ask questions.

\subsubsection{MMLU}
Massive Multitask Language Understanding (MMLU) \cite{hendrycks2021measuring} is another LLM benchmark, used for evaluating how well a model acquired knowledge during training. It uses 57 existing benchmarks, on subjects ranging from elementary to professional, and then averages these results to obtain the final score.

We decided to avoid running this benchmark, for the reason that the time complexity of this benchmark is large, and so instead we averaged the results over the remaining 3 benchmarks.

\section{Experiments}
Since Llama 2 is auto-regressive, it is hard to measure the average time reduction that we can expect from reducing the model size, as the size of the sequences can vary, especially for the task ``Hellaswag’’. Thus we will introduce a separate experiment for this, to test just how effective our proposed methods are at improving the inference costs.

\subsection{Preliminary Experiments}

As a preliminary, we checked with Llama 2 7b the effects of skipping the last 10\% of layers in comparison to the first 10\% of these layers. The results, shown below, allow us to exclude the need for checking skipping layers, as these empirically agree with the conclusion from Gao et al. \cite{DBLP:journals/corr/HuangSLSW16}. However, we encourage those with doubts to prove this for themselves.
 In the table below we show that skipping latter layers in Llama-2 7B (7B-90-Last) is better for performance than skipping the first 10\% of layers (7B-90-First)

\FloatBarrier
\begin{table}[h]
\caption{LLaMA-v2 preliminary First vs Last}
\label{tab:main_results}
\resizebox{0.93\textwidth}{!}{%
\begin{tabular}{@{}l|lllll@{}}
\toprule
\multicolumn{1}{c|}{\multirow{2}{*}{\textbf{Model}}} &  \multicolumn{5}{c}{\textbf{Performances}}     \\ 
\cmidrule(l){2-6} 
\multicolumn{1}{c|}{}                                & ARC & HellaSwag & TruthfulQA & MMLU & Average \\ \midrule
7B-90-Last\%   & 47.7 &  69.3 & 39.6 & - & 52.2 \\ \midrule

7B-90-First\%  & 28.7 & 26.3 & 47.9 & - & 34.3 \\

\bottomrule
\end{tabular}%
}
\end{table}
\FloatBarrier

To make sure our experiment was consistent, we conducted a dry run for the models, keeping them fully intact. We then compared the results we achieved with the results on the OpenLLM leaderboard \cite{open-llm-leaderboard}. We include these results in all of our tables in Chapter 4, as a comparison for accuracy.

For our third hypothesis, we are interested in whether high cosine similarities alone are enough of a criterion to skip certain layers. We would like to note that measured only the cosine similarities between the input and output of a full layer. We could have also done this for the sublayers, as those also have inputs and outputs of identical size, but we assume that if a full layer is redundant, then both the attention and feedforward sublayers have to be. Suppose one of the sublayers captures useful information, but then the opposite sublayer purposefully destroys the usefulness, making the whole layer redundant. If this is truly the case, then the pre-trained model updated its parameters in such a way to oppose the useful sublayer, which does not agree with classical gradient-based methods of optimization.

In the below figures we show the cosine similarities between layers for the three benchmark tasks we defined previously on the full Llama 2 7B model and the full Llama 2 13B model. The x-axis is the number of steps our recording software, WandB \cite{wandb}, used to track this data, and is related to time. The y-axis demonstrates the cosine similarity of each layer's inputs and outputs. We can see that these cosine similarities usually possess a strict hierarchy, and there is no change in ordinality as time goes on.

\begin{figure}[H]
    \centering
    \includegraphics{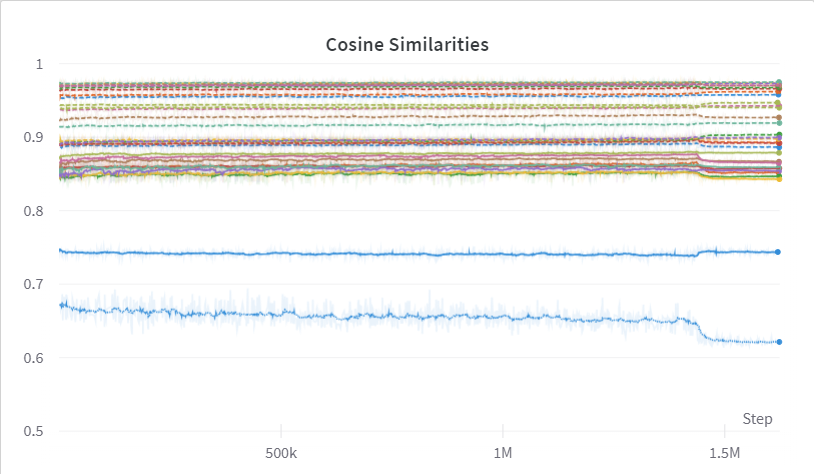}
    \caption{Cosine similarities between layers in Llama 2 7B}
    \label{fig:enter-label}
\end{figure}

In figure 3.1 we notice that the lowest cosine similarity is for the last layer which on average is 0.658. We provide these results in the appendix.

\begin{figure}[H]
    \centering
    \includegraphics{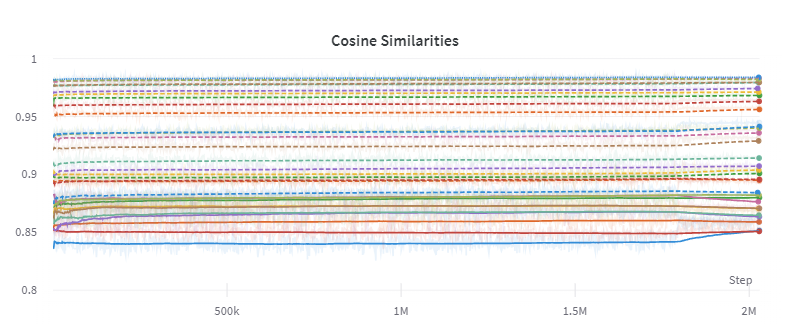}
    \caption{Cosine similarities between layers in Llama 2 13B}
    \label{fig:enter-label}
\end{figure}

Here we find that the cosine similarity for the last layer is much higher than for the 7B model, however it is still the lowest value from all values, with a similarity of 0.841. The full results will be provided in the appendix.

\subsection{Dropping full layers}

For our first experiment, we would like to examine the effects of skipping full layers in the LLM. To do this, we created a list of allowed layers for inference and skipped over the remaining disallowed layers. We then evaluate the models against the OpenLLM leaderboard \cite{open-llm-leaderboard} benchmarks \cite{clark2018think} \cite{zellers2019hellaswag} \cite{lin2022truthfulqa} as well as compare them to the fully intact models.
We skipped the last 10, 25, and 33 percent of layers, and then recorded the results for the benchmarks, These results were then aggregated (simple arithmetic mean) to obtain a final score for each skipped model.

We again mention that we will not evaluate MMLU in these experiments.

\subsection{Dropping sublayers}

For experiment 2, we would like to test how skipping different sublayers affects the performance of the model. Again we evaluated this by skipping the latter 10, 25, and 33 percent of the sublayers in each model. The results for each benchmark \cite{clark2018think} \cite{zellers2019hellaswag} \cite{lin2022truthfulqa}were then recorded and averaged to obtain the final score of the skipped model.

\subsection{Selective layer dropping}

Due to the preliminary experiments, we discovered that the lowest cosine similarity is between the ultimate and penultimate layer outputs. We hypothesize that this layer contains important information because of this disparity, while the layers before have very high cosine similarities, so that means not a lot of information is passed between these layers. Thus, we consider all the above experiments, but for the situation where the last layer is kept, and the skipped layers are moved down by one, so we keep the same number of layers skipped i.e. for a Llama 2 7B 10\% skipped, instead of 29 active layers, we have 28 layers plus the additional last layer, so the total number of layers skipped is the same.

\subsection{Measuring time-efficiency improvements}

Finally, we would like to test how reducing model size improves performance. To avoid having to encounter the autoregressive property of Llama 2, we create a new test.
We consider two collections of 1000 sequences, of 50 and 100 random tokens obtained from Llama 2 \cite{touvron2023llama2}. We will consider the same set of sequences for each model. 
The task for the model then is to predict the next token. We stop the autoregressiveness of Llama 2 by setting the maximum sequence length to 1, hence obtaining 1 token.
To find the time improvement, we track time from the beginning of the first sequence, to the end of the last sequence, and then averaging over the 1000 tokens predicted.
This reduces uncertainty due to only taking one-time measurements, in comparison to tracking time for each token separately.
We do not measure how accurate the model is in this scenario as there are no right answers, and we are purely interested in measuring the time efficiency of the models.

\section{Summary}

In this chapter, we covered the current research into optimizing large language models, as well as inspiring research for general deep learning networks. We then presented our proposed solution and how it differs from other methods. Further in that section we covered research that inspired us to look for such solutions.

Finally, we covered our experimental setup. We dissected the mechanisms we used for measuring performance, our preliminary research, and lastly, we discussed our experiments in detail, in preparation for the results we present in the chapter. 
\lhead{Results}
\chapter{Results}

In this chapter we will present our results that we obtained from running the experiments in the previous chapter. Our experiment aims were as follows:

\begin{itemize}

\item[-] Investigating dropping full layers

\item[-] Investigating the difference of dropping sublayers

\item[-] Investigating how last layers impact accuracy

\item[-] How time-efficient are our proposed solutions

\end{itemize}

We will cover each experiment result separately, restate the experiment setup, then presenting results. After stating the results, we then analyze them to obtain the optimal structure for the Llama 7B model for time-efficiency and accuracy.

\newpage

\section{Quantitative Results}

\subsection{Dropping full layers}
In the following table we show the results for dropping full layers for the Llama 2 7B model (7B) and the Llama 2 13B model (13B). We label the models with 0\%, 10\%, 25\%, 33\% of layers skipped as $\ast$-100\% $\ast$-90\%, $\ast$-75\%, $\ast$-66\% respectively. We present the results for the benchmarks as well as the average over those results. MMLU was excluded due to time constraints.
\FloatBarrier
\begin{table}[h]
\caption{LLaMA-v2 skipping full layer - results}
\label{tab:main_results}
\resizebox{0.93\textwidth}{!}{%
\begin{tabular}{@{}l|lllll@{}}
\toprule
\multicolumn{1}{c|}{\multirow{2}{*}{\textbf{Model}}} &  \multicolumn{5}{c}{\textbf{Performances}}     \\ 
\cmidrule(l){2-6} 
\multicolumn{1}{c|}{}                                & ARC & HellaSwag & TruthfulQA & MMLU & Average \\ \midrule
7B-66\%  & 35.2 & 46.8 & 46.2 & - & 42.7 \\
7B-75\%  & 38.3 & 53.0 & 45.1 & - & 45.5 \\
7B-90\%   & 47.7 &  69.3 & 39.6 & - & 52.2 \\
7B-100\%  & 53.1 & 78.6 & 38.8 & - & 54.3 \\ \midrule
13B-66\%  & 37.8 & 46.8 & 45.3 & - & 43.3 \\
13B-75\%  & 40.9 & 53.6 & 42.5 & - &  45.6 \\
13B-90\%  & 51.3 & 71.3 & 37.1 & - & 53.2 \\
13B-100\% & 59.6 & 82.1 & 36.9 & - & 59.5 \\
\bottomrule
\end{tabular}%
}
\end{table}
\FloatBarrier
From the above results, we notice that dropping whole layers are detrimental in large quantities, and dropping the last 10\% offer acceptable performance loss, of less than a few percentage points. What we also notice, is the TruthfulQA result does increase with less active layers, which matches with the proposition by \cite{shuster2021retrieval}\cite{lin2022truthfulqa}, where large models are less truthful.
\newpage
\subsection{Dropping sublayers(Attention vs. ffwd)}

\subsubsection{Attention sublayers}
In the following table we show the results for dropping attention sublayers for the Llama 2 7B model (7B) and the Llama 2 13B model (13B). We label the models with 0\%, 10\%, 25\%, 33\% of attention sublayers skipped as $\ast$-100\% $\ast$-90\%, $\ast$-75\%, $\ast$-66\% respectively. We present the results for the benchmarks as well as the average over those results. MMLU was excluded due to time constraints.
\begin{table}[h]
\caption{LLaMA-v2 skipping attention sublayers - results}
\label{tab:main_results}
\resizebox{0.93\textwidth}{!}{%
\begin{tabular}{@{}l|lllll@{}}
\toprule
\multicolumn{1}{c|}{\multirow{2}{*}{\textbf{Model}}} &  \multicolumn{5}{c}{\textbf{Performances}}     \\ 
\cmidrule(l){2-6} 
\multicolumn{1}{c|}{}                                & ARC & HellaSwag & TruthfulQA & MMLU & Average \\ \midrule
7B-66\%  & 51.2 & 77.0 & 42.2 & - &56.8 \\
7B-75\%  & 52.5 & 78.3 & 42.3 & - & 57.7\\
7B-90\%   & 52.8 &  78.9 & 40.0 & - & 57.2 \\
7B-100\%  & 53.1 & 78.6 & 38.8 & - & 54.3 \\ \midrule
13B-66\%  & 55.6 & 80.1 & 40.1 & - & 58.6 \\
13B-75\%  & 55.9 & 79.7 & 39.9 & - &  58.5 \\
13B-90\%  & 57.0 & 81.3 & 38.2 & - & 58.8 \\
13B-100\% & 59.6 & 82.1 & 36.9 & - & 59.5 \\
\bottomrule
\end{tabular}%
}
\end{table}
We observe that skipping attention sublayers, especially latter ones have very small impact on the performance, slightly decreasing the HellaSwag and ARC Challenge accuracy, but improving TruthfulQA. But overall, we notice that these changes are minimal, suggesting that skipping attention layers in LLMs, especially latter ones could become a valid compression technique, akin to \cite{venkataramanan2023skipattention} but for Visual Transformers. In fact, due to the improvement in TruthfulQA and insignificant changes to HellaSwag and ARC challenge, we observe that we actually improve on average benchmarks for Llama 2 7B model, and being competitive with Llama 2 13B.
\subsubsection{Feedforward sublayers}
In the following table we show the results for dropping feedforward (ffwd) sublayers for the Llama 2 7B model (7B) and the Llama 2 13B model (13B). We label the models with 0\%, 10\%, 25\%, 33\% of ffwd sublayers skipped as $\ast$-100\% $\ast$-90\%, $\ast$-75\%, $\ast$-66\% respectively. We present the results for the benchmarks as well as the average over those results. MMLU was excluded due to time constraints.
\begin{table}[h]
\caption{LLaMA-v2 skipping ffwd sublayers - results}
\label{tab:main_results}
\resizebox{0.93\textwidth}{!}{%
\begin{tabular}{@{}l|lllll@{}}
\toprule
\multicolumn{1}{c|}{\multirow{2}{*}{\textbf{Model}}} &  \multicolumn{5}{c}{\textbf{Performances}}     \\ 
\cmidrule(l){2-6} 
\multicolumn{1}{c|}{}                                & ARC & HellaSwag & TruthfulQA & MMLU & Average \\ \midrule
7B-66\%  & 35.1 & 52.5 & 42.2 & - & 43.3 \\
7B-75\%  & 40.4 & 60.3 & 39.2 & - & 46.6\\
7B-90\%  & 48.5 & 71.4 & 38.0 & - & 52.6 \\
7B-100\% & 53.1 & 78.6 & 38.8 & - & 54.3 \\ \midrule
13B-66\%  & 41.6 & 56.9 & 40.7 & - & 46.4 \\
13B-75\%  & 47.3 & 65.2 & 40.0 & - &  50.3 \\
13B-90\%  & 54.2 & 75.8 & 38.3 & - & 56.1 \\
13B-100\% & 59.6 & 82.1 & 36.9 & - & 59.5 \\
\bottomrule
\end{tabular}%
}
\end{table}
Feedforward sublayers are more impactful on performance. We can see that even a skipping the last 10\% of sublayers severely impacts accuracy when comparing this result to the attention sublayers. Combining this with \cite{zhang2022layers}, we can conclude that robustness of a layer (or sublayer) correlates with how ``skippable’’ that layer is. 
\newpage
\subsection{Last layer inclusion}
Below we show three tables of results for including the last layer in each model. We show the results for full layer dropping, attention sublayer and ffwd sublayers as well. We shift our dropped layers by one, so we are still skipping the same amount of layers as before, but now with the last layer at the end.
In these tables we label the models with 0\%, 10\%, 25\%, 33\% of layers or sublayers skipped as $\ast$-100\% $\ast$-90\%, $\ast$-75\%, $\ast$-66\% respectively, for each base model Llama 2 7B (7B) or Llama 2 13B (13B). We display the results for individual benchmarks as well as their average.
\begin{table}[h]
\caption{LLaMA-v2 skip full layers with last layer - results}
\label{tab:main_results}
\resizebox{0.93\textwidth}{!}{%
\begin{tabular}{@{}l|lllll@{}}
\toprule
\multicolumn{1}{c|}{\multirow{2}{*}{\textbf{Model}}} &  \multicolumn{5}{c}{\textbf{Performances}}     \\ 
\cmidrule(l){2-6} 
\multicolumn{1}{c|}{}                                & ARC & HellaSwag & TruthfulQA & MMLU & Average \\ \midrule
7B-66\%  & 32.0 & 45.8 & 46.9 & - & 43.3 \\
7B-75\%  & 34.5 & 49.4 & 45.9 & - & 46.6\\
7B-90\%  & 46.5 & 73.1 & 41.8 & - & 52.6 \\
7B-100\% & 53.1 & 78.6 & 38.8 & - & 54.3 \\ \midrule
13B-66\%  & 35.1 & 50.0 & 46.9 & - & 46.4 \\
13B-75\%  & 38.7 & 56.6 & 43.7 & - & 50.3 \\
13B-90\%  & 51.2 & 78.1 & 38.0 & - & 56.1 \\
13B-100\% & 59.6 & 82.1 & 36.9 & - & 59.5 \\
\bottomrule
\end{tabular}%
}
\end{table}

\begin{table}[h]
\caption{LLaMA-v2 skip attention sublayers with last layer - results}
\label{tab:main_results}
\resizebox{0.93\textwidth}{!}{%
\begin{tabular}{@{}l|lllll@{}}
\toprule
\multicolumn{1}{c|}{\multirow{2}{*}{\textbf{Model}}} &  \multicolumn{5}{c}{\textbf{Performances}}     \\ 
\cmidrule(l){2-6} 
\multicolumn{1}{c|}{}                                & ARC & HellaSwag & TruthfulQA & MMLU & Average \\ \midrule
7B-66\%  & 49.3 & 77.1 & 40.5 & - & 55.6 \\
7B-75\%  & 51.8 & 78.3 & 41.1 & - & 57.1\\
7B-90\%   & 51.9 &  78.7 & 39.4 & - & 56.7 \\
7B-100\%  & 53.1 & 78.6 & 38.8 & - & 54.3 \\ \midrule
13B-66\%  & 56.8 & 82.1 & 38.0 & - & 59.0 \\
13B-75\%  & 57.5 & 82.1 & 37.0 & - &  58.9 \\
13B-90\%  & 58.9 & 82.4 & 36.6 & - & 59.3 \\
13B-100\% & 59.6 & 82.1 & 36.9 & - & 59.5 \\
\bottomrule
\end{tabular}%
}
\end{table}

\begin{table}[h]
\caption{LLaMA-v2 skip ffwd sublayers with last layer - results}
\label{tab:main_results}
\resizebox{0.93\textwidth}{!}{%
\begin{tabular}{@{}l|llll|l@{}}
\toprule
\multicolumn{1}{c|}{\multirow{2}{*}{\textbf{Model}}} &  \multicolumn{5}{c}{\textbf{Performances}}     \\ 
\cmidrule(l){2-6} 
\multicolumn{1}{c|}{} & ARC & HellaSwag & TruthfulQA & MMLU & Average \\ \midrule
7B-66\%  & 32.0 & 45.8 & 46.9 & - &  41.6\\
7B-75\%  & 34.5 & 49.4 & 45.9 & - & 43.3\\
7B-90\%   & 46.5 &  73.1 & 41.8 & - & 53.8 \\
7B-100\%  & 53.1 & 78.6 & 38.8 & - & 54.3 \\ \midrule
13B-66\%  & 35.1 & 50.0 & 46.9 & - & 44.0 \\
13B-75\%  & 38.7 & 56.6 & 43.7 & - & 46.3 \\
13B-90\%  & 51.2 & 78.1 & 38.0 & - & 55.8 \\
13B-100\% & 59.6 & 82.1 & 36.9 & - & 59.5 \\
\bottomrule
\end{tabular}%
}
\end{table}
\FloatBarrier

Surprisingly, we notice that skipping layers except the latter layers reduces performances for more layers skipped, except for skipping the attention layers. This is even more exaggerated compared to just dropping layers. The reason for this could be again attributed to the (lack of) robustness of feedforward sublayers, as the last layer now has to process perturbed information from earlier layers, which did not occur when just dropping layers.
\newpage
\subsection{Time-efficiency of Llama 7B}

To measure the efficiency of the networks we conducted a separate experiment, where we record the time it takes for the model to output a sequence of length 1, averaging over 1000 sequences. We conducted this experiment for both 50 and 100 length input sequences. We notice that full layer droppings do improve time costs the best, followed by attention sublayers, and then feedforward sublayers which do not impact the speed of processing a lot.

We report the time$\times 10^2$ (for clarity)it takes to predict 1 token for 1000 sequences  as well as the percentage improvement. We show the results of this experiment for Llama 2 7B with 0\%, 10\%, 25\%, 33\% of layers skipped and we label these as 7B-100\%, 7B-90\%, 7B-75\%, 7B-66\% respectively.

\begin{table}[h]
\caption{LLaMA-v2 time results, 50 length sequence, no last layer}
\label{tab:main_results}
\resizebox{\textwidth}{!}{%
\begin{tabular}{@{}l|ll|ll|ll@{}}
\toprule
\multicolumn{1}{c|}{\multirow{2}{*}{\textbf{Model}}} &  \multicolumn{2}{c}{\textbf{Full}} & \multicolumn{2}{c}{\textbf{Attention}} & \multicolumn{2}{c}{\textbf{ffwd}}   \\ 
\cmidrule(l){2-7}
\multicolumn{1}{c|}{} & Time(s) $\times 10^2$ & (\%) & Time(s) $\times 10^2$ & (\%) & Time(s) $\times 10^2$ & (\%)\\ \midrule
7B-66\%  & 31.35 & 32.96 & 36.72 & 21.47 &  43.51 & 6.95 \\
7B-75\%  & 35.48 & 24.12 & 39.46 & 15.61 & 42.88 & 8.30\\
7B-90\%   & 43.31 & 7.38 & 42.93 & 8.19 & 44.17 & 5.53 \\
\midrule
7B-100\%  & 46.76 & 0 & - & - & - & - \\ 
\bottomrule
\end{tabular}%
}
\end{table}
\FloatBarrier
\begin{table}[h]
\caption{LLaMA-v2 time results, 50 length sequence, last layer included}
\label{tab:main_results}
\resizebox{\textwidth}{!}{%
\begin{tabular}{@{}l|ll|ll|ll@{}}
\toprule
\multicolumn{1}{c|}{\multirow{2}{*}{\textbf{Model}}} &  \multicolumn{2}{c}{\textbf{Full}} & \multicolumn{2}{c}{\textbf{Attention}} & \multicolumn{2}{c}{\textbf{ffwd}}   \\ 
\cmidrule(l){2-7}
\multicolumn{1}{c|}{} & Time(s) $\times 10^2$ & (\%) & Time(s) $\times 10^2$ & (\%) & Time(s) $\times 10^2$ & (\%)\\ \midrule
7B-66\%  & 31.78 & 32.04 & 36.92 & 21.04 &  41.31 & 11.66\\
7B-75\%  & 34.98 & 25.19 & 40.24 & 13.94 & 42.62 & 8.85\\
7B-90\%   & 40.92 & 12.49 & 42.43 & 9.26 & 43.51 & 6.95  \\
\midrule
7B-100\%  & 46.76 & 0 & - & - & - & - \\ 
\bottomrule
\end{tabular}%
}
\end{table}
\FloatBarrier

\begin{table}[h]
\caption{LLaMA-v2 time results, 100 length sequence, no last layer}
\label{tab:main_results}
\resizebox{\textwidth}{!}{%
\begin{tabular}{@{}l|ll|ll|ll@{}}
\toprule
\multicolumn{1}{c|}{\multirow{2}{*}{\textbf{Model}}} &  \multicolumn{2}{c}{\textbf{Full}} & \multicolumn{2}{c}{\textbf{Attention}} & \multicolumn{2}{c}{\textbf{ffwd}}   \\ 
\cmidrule(l){2-7}
\multicolumn{1}{c|}{} & Time(s) $\times 10^2$ & (\%) & Time(s) $\times 10^2$ & (\%) & Time(s) $\times 10^2$ & (\%)\\ \midrule
7B-66\%  & 32.36 & 32.58 & 38.97 & 18.18 &  43.08 & 10.25\\
7B-75\%  & 36.58 & 23.79 & 41.27 & 14.02 & 44.13 & 8.06 \\
7B-90\%   & 43.65 & 9.06 & 44.62 & 7.04 & 46.30 & 3.54 \\
\midrule
7B-100\%  & 48.00 & 0 & - & - & - & - \\ 
\bottomrule
\end{tabular}%
}
\end{table}
\FloatBarrier
\begin{table}[h]
\caption{LLaMA-v2 time results, 100 length sequence, last layer included}
\label{tab:main_results}
\resizebox{\textwidth}{!}{%
\begin{tabular}{@{}l|ll|ll|ll@{}}
\toprule
\multicolumn{1}{c|}{\multirow{2}{*}{\textbf{Model}}} &  \multicolumn{2}{c}{\textbf{Full}} & \multicolumn{2}{c}{\textbf{Attention}} & \multicolumn{2}{c}{\textbf{ffwd}}   \\ 
\cmidrule(l){2-7}
\multicolumn{1}{c|}{} & Time(s) $\times 10^2$ & (\%) & Time(s) $\times 10^2$ & (\%) & Time(s) $\times 10^2$ & (\%)\\ \midrule
7B-66\%  & 32.05 &33.23  & 38.52 & 19.75 &  42.66 & 11.13\\
7B-75\%  & 36.41 & 24.15 & 41.00 & 14.58 & 43.92 & 8.50\\
7B-90\%   & 43.28 & 9.83 & 44.27 & 7.77 & 45.20 & 5.83 \\
\midrule
7B-100\%  & 48.00 & 0 & - & - & - & - \\ 
\bottomrule
\end{tabular}%
}
\end{table}
\FloatBarrier

As covered in Chapter 3, we show this only for the 7B as we run out of memory when performing this on the 13B model (a quantized model takes $\sim$ 39 GB, and evaluating a sequence of 100 tokens, maxes out the total possible  space available on an NVIDIA A100 40GB GPU).

We notice that deciding to include the last layer or not has no significant impact on the time improvement of the models, as we expected, since we skip the same amount of layers. We also note that increasing sequence length by 50 has no significant impact on the percentage improvement, but does increase the base inference costs.

In the table below we show the runtimes of our experiments for each model, Llama 2 7B and Llama 2 13B, as well as for every layer skipping experiment, where 100\% refers to a model with no skipped layers, and 66\% refers to a model with 33\% of layers skipped. We show the run times for full layer dropping, as well as attention and feedforward (ffwd) sublayer dropping.
\begin{table}[h]
\caption{LLaMA-v2 run times for our experiments}
\label{tab:main_results}
\resizebox{0.93\textwidth}{!}{%
\begin{tabular}{@{}l|lll|lll@{}}
\toprule
\multicolumn{1}{c|}{\multirow{2}{*}{\textbf{Model}}} &  \multicolumn{3}{c}{\textbf{No last layer}} & \multicolumn{3}{c}{\textbf{Last layer included}}     \\ 
\cmidrule(l){2-7} 
\multicolumn{1}{c|}{} & Full & Attention & ffwd & Full & Attention & ffwd \\ \midrule
7B-66\%  & 9h 20m & 12h 43m & 13h 19m & 11h 27m &  12h 38m & 5h 15m\\
7B-75\%  & 11h 37m & 13h 2m & 13h 27m & 12h 15m & 13h 0m & 5h 27m\\
7B-90\%   & 12h 45m &  13h 22m & 13h 44m & 12h 55m & 13h 29m & 7h 36m \\
7B-100\%  & 13h 13m & - & - & - & - & - \\ \midrule
13B-66\%  & 13h 1m & 13h 20m & 13h 45m & 11h 42m & 13h 33m & 4h 43m\\
13B-75\%  & 13h 31m & 13h 49m & 14h 5m & 12h 40m & 13h 50m & 9h 22m\\
13B-90\%  & 13h 22m & 14h 25m & 14h 23m & 13h 22m & 14h 1m & 7h 57m\\
13B-100\% & 13h 59m & - & - & - & - & - \\
\bottomrule
\end{tabular}%
}
\end{table}
\FloatBarrier
These runtimes may not representative of how a model increases speed, as the answers provided by the models can vary in length, however it could show how some size reductions could impact average sequence length. This is especially visible with how models with less layers take more time to run. These runtimes could have been affected by many factors, during our experiments, so we will not consider these as valid results, however we decide to include them anyway.

\section{Qualitative Analysis}

From the results above, we can conclude that the best course of action for large language model compression, is by skipping attention layers. We notice that unlike with full layer dropping, latter attention sublayers are redundant and therefore by skipping those sublayers we can keep the level of performance close to the full model, and even improve the performance of LLMs, due to the improvement of truthfullness. We notice that the feedforward sublayers are critical for inference, more so than their attention sublayer counterparts, and skipping them, either separately or dropping them as full layers, results in significant performance drops. They also do not impact the time of inference as much as attention sublayers. We notice that skipping attention layers does not necessarily improve time efficiency for longer sequences, even though this was expected according to Vaswani et al. \cite{DBLP:journals/corr/VaswaniSPUJGKP17}, where attention layers were defined to have a quadratic complexity with respect to the sequence length. In our experiment, however, the percentage change in improvement is not large enough to agree with this defined complexity.

\lhead{Conclusion}
\chapter{Conclusion}

From our experiments, we empirically show that, within the considered benchmarks, latter attention layers have no significant impact on performance, and can even be a disadvantage in some tasks. This conclusion agrees with the research done by \cite{venkataramanan2023skipattention}, where a similar result was found for Visual Transformers, however this phenomenon is still surprising, considering all of the weight were used during the training of the model. Our research could inspire the training of LLMs with no attention sublayers in the tails, as we empirically show these are redundant for inference, and could mean that there is no point even training these weights in the first place.

 Attention sublayers are also heavy computationally, as they take up a lot of memory when loading onto GPUs. These drawbacks of attention sublayers, added with the fact that they provide almost no information on inference, means attentionless LLMs might become more common. We also would like to note that any small drops in performance of skipping attention layers, could be addressed to the sensitive and non-robustness of the MLP layers, where even minor perturbations could affect their outputs significantly. This could be addressed with fine-tuning; completely negating any possible negative effects of skipping attention sublayers.

We also mention that cosine similarities can be used to decide when to include a layer and when not to remove it, as the results for skipping 10\% of layers did improve when we kept the last layer, however we suggest that skipping large quantities of layers could reduce performance more by keeping the last layer, due to the sensitivity of the feedforward sublayers. This method however has potential, and implementing a periodic skipping, i.e. skipping every other layer could be reduce the perturbations within the inputs in between layers, thus reducing the negative performance of including the last layer.

Our research was done on the latest Llama 2 models for sizes of 7B and 13B, so it makes sense to propose further research into other models and sizes, to explore how much can we really skip in behemoths such as GPT4 or at least GPT3.5. These methods could be interesting in other types of LLMs such as chatbots or instruct bots, both of which are fine-tuned to be better for their selected tasks.

Another interesting research direction could be to fine-tune this reduced model, to see how much potential these models possess, since LLM models are not trained to convergence due to their size, such a compression could make it feasible, especially due to the amount of parameters in attention sublayers.

In conclusion, our research shows that Large Language Models can be compressed in a rather simplistic method by removing attention sublayers in latter parts of the model, as these are computationally heavy and redundant, and sometimes even hinder performance. We empirically show that for some models we can reduce time costs for generating one token by 21\% whilst also keeping the accuracy of the original model, as well as even improving on this. 




\addtocontents{toc}{\vspace{2em}} 

\appendix 

\chapter{Appendix A: Llama 7B Cosine similarities}
\begin{verbatim}
    "layer 1 cos_similarity": 0.7431640625,
    "layer 2 cos_similarity": 0.8505859375,
  "layer 3 cos_similarity": 0.857421875,
  "layer 4 cos_similarity": 0.8466796875,
  "layer 5 cos_similarity": 0.841796875,
  "layer 6 cos_similarity": 0.85302734375,
  "layer 7 cos_similarity": 0.85791015625,
  "layer 8 cos_similarity": 0.86572265625,
  "layer 9 cos_similarity": 0.86328125,
  "layer 10 cos_similarity": 0.87841796875,
  "layer 11 cos_similarity": 0.8857421875,
  "layer 12 cos_similarity": 0.89208984375,
  "layer 13 cos_similarity": 0.8916015625,
  "layer 14 cos_similarity": 0.90380859375,
  "layer 15 cos_similarity": 0.8974609375,
  "layer 16 cos_similarity": 0.89892578125,
  "layer 17 cos_similarity": 0.91943359375,
  "layer 18 cos_similarity": 0.9267578125,
  "layer 19 cos_similarity": 0.9423828125,
  "layer 20 cos_similarity": 0.947265625,
  "layer 21 cos_similarity": 0.95751953125,
  "layer 22 cos_similarity": 0.9619140625,
  "layer 23 cos_similarity": 0.9658203125,
  "layer 24 cos_similarity": 0.96728515625,
  "layer 25 cos_similarity": 0.97412109375,
  "layer 26 cos_similarity": 0.97314453125,
  "layer 27 cos_similarity": 0.97509765625,
  "layer 28 cos_similarity": 0.97021484375,
  "layer 29 cos_similarity": 0.970703125,
  "layer 30 cos_similarity": 0.939453125,
  "layer 31 cos_similarity": 0.65832890625,
\end{verbatim}

\chapter{Appendix B: Llama 2 13B Cosine Similarities}\begin{verbatim}
"layer 1 cos_similarity": 0.85986328125,
 "layer 2 cos_similarity": 0.857421875,
  "layer 3 cos_similarity": 0.8525390625,
  "layer 4 cos_similarity": 0.8798828125,
  "layer 5 cos_similarity": 0.86865234375,
  "layer 6 cos_similarity": 0.85986328125,
  "layer 7 cos_similarity": 0.86181640625,
  "layer 8 cos_similarity": 0.8671875,
  "layer 9 cos_similarity": 0.87158203125,
  "layer 10 cos_similarity": 0.88134765625,
  "layer 11 cos_similarity": 0.88232421875,
  "layer 12 cos_similarity": 0.8935546875,
  "layer 13 cos_similarity": 0.89404296875,
  "layer 14 cos_similarity": 0.90234375,
  "layer 15 cos_similarity": 0.9052734375,
  "layer 16 cos_similarity": 0.9072265625,
  "layer 17 cos_similarity": 0.9150390625,
  "layer 18 cos_similarity": 0.93212890625,
  "layer 19 cos_similarity": 0.9384765625,
  "layer 20 cos_similarity": 0.94189453125,
  "layer 21 cos_similarity": 0.94482421875,
  "layer 22 cos_similarity": 0.95751953125,
  "layer 23 cos_similarity": 0.96435546875,
  "layer 24 cos_similarity": 0.96875,
  "layer 25 cos_similarity": 0.9716796875,
  "layer 26 cos_similarity": 0.97509765625,
  "layer 27 cos_similarity": 0.97998046875,
  "layer 28 cos_similarity": 0.97998046875,
  "layer 29 cos_similarity": 0.9833984375,
  "layer 30 cos_similarity": 0.98193359375,
  "layer 31 cos_similarity": 0.98388671875,
  "layer 32 cos_similarity": 0.98291015625,
  "layer 33 cos_similarity": 0.9833984375,
  "layer 34 cos_similarity": 0.982421875,
  "layer 35 cos_similarity": 0.982421875,
  "layer 36 cos_similarity": 0.9775390625,
  "layer 37 cos_similarity": 0.97216796875,
  "layer 38 cos_similarity": 0.9560546875,
  "layer 39 cos_similarity": 0.8408203125

\end{verbatim} 

\addtocontents{toc}{\vspace{2em}}  
\backmatter

\label{Bibliography}
\lhead{\emph{Bibliography}}  
\bibliographystyle{unsrtnat}  

\end{document}